\documentclass{article}

\PassOptionsToPackage{numbers, compress}{natbib}
\usepackage[preprint]{neurips_2026}
\usepackage{amsmath}

\usepackage[utf8]{inputenc} 
\usepackage[T1]{fontenc}    
\usepackage{hyperref}       
\usepackage{url}            
\usepackage{booktabs}       
\usepackage{amsfonts}       
\usepackage{nicefrac}       
\usepackage{microtype}      
\usepackage{xcolor}  
\usepackage[table]{xcolor}
\usepackage{tabularx}
\usepackage{graphicx}
\usepackage{array}
\usepackage{tabularx}
\usepackage{multirow}
\usepackage{pifont}
\usepackage{wrapfig}
\usepackage{float}
\definecolor{sectiongray}{RGB}{245,245,245}
\definecolor{gain}{RGB}{220,217,246}
\definecolor{sota}{RGB}{239,238,250}
\newcolumntype{C}{>{\centering\arraybackslash}p{1.5cm}}
\newcolumntype{Y}{>{\raggedright\arraybackslash}X}

\title{EasyLens: A Training-Free Plug-and-Play Subtle-Lesion Representation Amplifier for Medical Vision-Language Models}

\author{%
  \textbf{Hao Wang$^{4,2,1*}$ \quad
  Qiwei Zeng$^{1,*}$ \quad
  Jinghao Lin$^{3,*}$ \quad
  Shuchang Ye$^{2}$} \\
  \textbf{Yuezhe Yang$^{4}$ \quad
  Yige Peng$^{4}$ \quad
  Haoyuan Che$^{1,\dagger}$ \quad
  Jinman Kim$^{2,\dagger}$ \quad
  Lei Bi$^{4,\dagger}$} \\
  \normalfont
  $^{1}$Jilin University, Changchun, China \\
  $^{2}$School of Computer Science, The University of Sydney, Sydney, NSW, Australia \\
  $^{3}$Northeastern University, Shenyang, China \\
  $^{4}$Institute of Translational Medicine, Shanghai Jiao Tong University, Shanghai, China \\
  $^{*}$Equal contribution. \\
  $^{\dagger}$Corresponding authors\\
  \texttt{chy@jlu.edu.cn} \quad
  \texttt{jinman.kim@sydney.edu.au} \quad
  \texttt{lei.bi@sjtu.edu.cn}
}

\begin{document}

\maketitle

\begin{abstract}
Medical vision-language models (VLMs) have shown increasing potential for clinical image interpretation, including lesion detection and report generation. However, their practical utility remains limited by insufficient sensitivity to subtle lesions, whose visual evidence is often sparse, low-contrast, and embedded within complex anatomical context. As local visual tokens are aggregated, these weak lesion cues can become underrepresented in global image representations, making them difficult for medical VLMs to recognize. Existing efforts to improve lesion sensitivity mainly rely on medical-domain vision-encoder pre-training, clinical-term-guided alignment, or trainable pathological representation enhancement. Although effective, these approaches usually require additional training or model-specific adaptation and may overfit to particular disease morphologies, limiting their applicability to frozen medical VLMs. To address these limitations, we propose \textbf{EasyLens}, a training-free plug-and-play subtle-lesion representation amplifier for medical VLMs. EasyLens first constructs \textbf{EasyBank}, a pathology-anatomy prototype space that provides lesion-related prototypes and anatomy-aware normal references for comparing suspicious patches against both pathological and normal anatomical patterns. To avoid blindly amplifying normal tissues, \textbf{EasyTag} selects lesion-relevant patches through counterfactual prototype reasoning. To counteract the dilution of subtle lesion cues in global image representations, \textbf{EasyAmplifier} strengthens the selected lesion-relevant patch representations through morphology-guided residual enhancement, thereby increasing their contribution to the global image embedding. Experiments on multiple medical image datasets and frozen medical VLM backbones show that EasyLens consistently improves subtle-lesion detection and outperforms existing encoder-enhancement baselines without model fine-tuning. Code is available at: https://github.com/wanghao1118/EasyLens
\end{abstract}

\section{Introduction}
Medical vision-language models (VLMs) are increasingly explored for clinical image interpretation, including lesion detection and report generation \cite{hartsock2024vision,royer2024multimedeval}. As diagnostic decision-support tools, their practical value depends on reliable recognition across a broad spectrum of lesion appearances, rather than only on highly visible abnormalities \cite{cheng2025understanding}. Here, salient lesions refer to abnormalities with large spatial extent, high contrast, or pronounced morphological signatures, whereas subtle lesions exhibit sparse, low-contrast, or weakly distinguishable visual cues embedded within complex anatomical context \cite{li2024anatomical,mao2020abnormality}. Although salient lesions can often be captured by strong visual patterns, subtle lesions provide weak local evidence that can be easily confused with normal anatomical variation. As local visual tokens are aggregated into global image representations, these weak lesion cues may become underrepresented, causing medical VLMs to miss or under-recognize subtle abnormalities \cite{tivnan2023task}. Therefore, improving the sensitivity of medical VLMs to subtle lesions is critical for making them more reliable in clinical image interpretation \cite{zhang2024single}.

\begin{figure}[t]
    \centering
    \begin{minipage}{\linewidth}
        \centering
        \includegraphics[width=\linewidth]{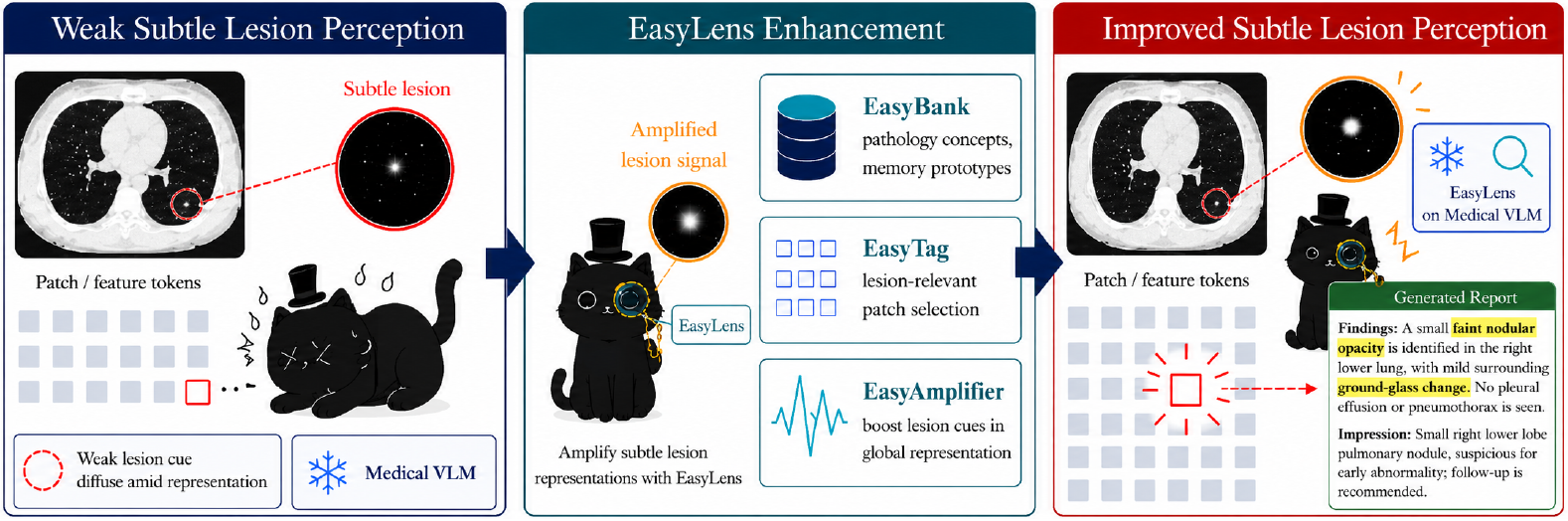}
        \vspace{-15pt}
        \caption{\textbf{EasyLens} brings weak subtle-lesion cues into focus in frozen medical VLMs.}
        \label{fig:overview}
    \end{minipage}
    \vspace{-15pt}
\end{figure}

To improve the image interpretation capabilities of medical VLMs, existing approaches have primarily focused on enhancing their visual encoders to extract richer semantic features from medical images and better align image representations with clinical text \cite{zhang2022contrastive,boecking2022making}. Initially, many approaches pre-trained vision encoders to bridge the domain gap on large-scale paired medical images and reports \cite{zhang2022contrastive}, enabling them to capture pathological relevant semantic information when encoding medical images \cite{wang2022medclip}. Although these pre-trained encoders effectively improve the interpretation capability of medical VLMs in the disease diagnosis, these coarse-grained building methods for pre-trained datasets make vision encoders lack the effective guidance to detect subtle lesions \cite{mao2020abnormality,li2024anatomical}.

To provide clinically relevant guidance and enhance the pathological semantics of medical image embeddings, subsequent studies introduced domain-specific modules into pretrained vision encoders in medical VLMs. To guide visual encoding with clinical semantics, several studies explicitly inject clinical semantic entities into the image encoding process. MedKLIP \cite{wu2023medklip} extracts disease-related clinical entities and their spatial attributes from radiology reports to establish entity-patch correspondence, while KAD \cite{zhang2023knowledge} constructs clinical entities and relations from reports and uses knowledge-guided disease queries to condition visual representation learning. Complementary to clinical-entity guidance, other studies enhance pathological semantics by refining patch-level visual representations. MLIP \cite{liu2024mlip} improves local image-text alignment through masked local representation learning, while AdaMatch \cite{chen2024fine} uses adaptive patch matching to capture abnormalities with varying sizes and locations. PLACE \cite{wang2025improving} further enriches fine-grained visual details through pathological-level alignment and patch correlation modeling.

Although these encoder-enhancement methods improve the detection of subtle lesions, their general applicability to current medical VLMs remains limited. First, clinical-semantic guidance is insufficient to cover the complex pathological variations in real clinical scenarios. Since lesions often present with diverse morphological patterns and imaging appearances, supervision based on predefined clinical semantics provides coarse guidance and fails to capture subtle pathological details, leading to confusion among visually similar abnormalities. Specifically, subtle lesions occupy a limited portion of image patches, making fine-grained pathological detail preservation essential for accurate detection. Second, representation-level enhancement typically relies on extensive training or re-training, leading to considerable computational and time costs. In addition, adapting the encoder to specific pathological patterns reduce its sensitivity to other abnormalities, thereby limiting the applicability of medical VLMs across diverse clinical scenarios.

To address these limitations, we propose \textbf{EasyLens}, a training-free plug-and-play subtle-lesion representation amplifier for medical VLMs. EasyLens is designed to strengthen weak lesion cues that are preserved in frozen visual representations but become underrepresented during global image aggregation. It first constructs \textbf{EasyBank}, a pathology-anatomy prototype space that provides lesion-related prototypes and anatomy-aware normal references for patch-level comparison. By contrasting suspicious patches with both pathological prototypes and normal anatomical references, EasyBank supports fine-grained discrimination between subtle lesion evidence and normal anatomical variation. Built upon EasyBank, \textbf{EasyTag} selects lesion-relevant patches through counterfactual prototype reasoning, thereby avoiding blind amplification of normal tissues. \textbf{EasyAmplifier} then strengthens the selected lesion-relevant patch representations through morphology-guided residual enhancement, increasing their contribution to the global image embedding while preserving the original visual context. Both modules operate without updating model parameters or requiring lesion annotations at inference time, making EasyLens applicable to frozen medical VLMs. We validate EasyLens on a unified subtle-lesion benchmark built from ReXGroundingCT, LIDC-IDRI, and AbdomenAtlas 3.0 Mini. Experiments across multiple frozen medical VLM backbones show that EasyLens consistently improves subtle-lesion detection and report generation, and outperforms existing encoder-enhancement baselines without model fine-tuning. Our contributions are summarized as follows:

\noindent \textbf{(1)} We propose \textbf{EasyLens}, a training-free plug-and-play amplifier that improves subtle-lesion recognition in frozen medical VLMs by exploiting latent pathological evidence through prototype-based reasoning.

\noindent \textbf{(2)} We construct \textbf{EasyBank}, a pathology-anatomy prototype space that organizes lesion-related prototypes and anatomy-aware normal references, and design \textbf{EasyTag}, a counterfactual prototype-guided patch selector that identifies lesion-relevant regions through fine-grained pathological comparison.

\noindent \textbf{(3)} We introduce \textbf{EasyAmplifier}, a morphology-guided residual semantic amplifier that enhances disease-related morphological semantics in selected patch representations without model fine-tuning or inference-time lesion annotations.

\section{Related Work}
\subsection{Medical VLMs in Radiology}

Recent medical VLMs in radiology have evolved from narrow image-to-report pipelines into more general systems for multi-task interpretation, interactive querying, visual grounding, and structured reasoning. On 2D chest radiographs, recent studies have improved clinical usability by introducing agentic tool use, anatomy-centric reasoning, fine-grained vision-language alignment, and pixel-grounded interaction. For example, MedRAX integrates multimodal tools and large models for complex chest X-ray interpretation \cite{pmlr-v267-fallahpour25a}, AOR performs anatomy-centric region-level reasoning \cite{li2025aor}, and RadZero strengthens fine-grained alignment for zero-shot classification, grounding, and segmentation \cite{park2025radzero}. MIMO further extends medical VLMs beyond text-only responses by supporting visual referring inputs and pixel-grounded outputs \cite{chen2025mimo}. These works reflect a shift from holistic report generation toward more interactive and evidence-grounded radiological interpretation.

Another major trend extends radiology VLMs from 2D radiographs to volumetric understanding. Argus studies large-scale 3D CT report generation and highlights the importance of vision encoder pretraining, visual token compression, and model/data scaling for high-resolution 3D radiology VLMs \cite{liu2025argus}. BTB3D further shows that effective volumetric tokenization is more critical than simply enlarging language backbones for scalable 3D medical VLMs \cite{hamamci2025better}. Recent public systems continue this direction through organ-separated CT-language modeling, variable-length 3D visual tokens, and native CT/MRI interpretation \cite{yamamoto2026totalfm,fang2026photon,sellergren2025medgemma}. In parallel, clinically grounded radiology VLMs incorporate step-by-step verification, reinforcement learning, workflow-level context, or radiologist gaze to align model reasoning with expert diagnostic procedures \cite{fan2025chestx,zhang2026reasoning,liu2026seeing,lee2026seeing}. Despite these advances, most radiology VLMs still primarily optimize global image-report alignment, report generation, VQA, or coarse region-level grounding. Sparse subtle-lesion cues can therefore be weakened by holistic diagnostic semantics or compressed visual tokens, leaving a gap between strong global radiology understanding and robust perception of subtle lesion-carrying patches.

\subsection{Subtle Lesion Detection in Medical VLMs}

Subtle lesions remain challenging for medical VLMs because their diagnostic cues are weak, spatially sparse, and often overwhelmed by surrounding anatomical structures. To improve the perception of such lesions, existing studies have enhanced medical visual representations by injecting clinically relevant guidance into pretrained vision encoders. MedKLIP \cite{wu2023medklip} extracts disease-related clinical entities and their spatial attributes from radiology reports to establish entity-patch correspondence, while KAD \cite{zhang2023knowledge} constructs clinical entities and relations from reports and uses knowledge-guided disease queries to condition visual representation learning. Beyond clinical-entity guidance, other studies refine patch-level pathological semantics to better capture local abnormalities. MLIP \cite{liu2024mlip} improves local image-text alignment through masked local representation learning, AdaMatch \cite{chen2024fine} uses adaptive patch matching to capture abnormalities with varying sizes and locations, and PLACE \cite{wang2025improving} further enriches fine-grained visual details through pathological-level alignment and patch correlation modeling.

Another related direction strengthens subtle lesion perception through region-aware grounding and abnormality-sensitive tuning. VividMed \cite{luo2025vividmed} and MIMO \cite{chen2025mimo} extend medical VLMs with segmentation, referring, and pixel-grounded outputs, enabling pathological findings to be associated with specific image regions. Reg2RG \cite{chen2025large} incorporates region-guided referring and grounding into CT report generation, while UMed-LVLM \cite{zhou2025improving} and MMedPO\cite{zhu2024mmedpo} improve abnormal-region sensitivity through abnormal-aware fine-tuning or clinical-aware preference optimization. These studies push medical VLMs from coarse image-level diagnosis toward more localized subtle lesion understanding. However, most existing methods rely on predefined clinical semantics, explicit grounding modules, region-level supervision, or additional fine-tuning, making them costly and less directly applicable to frozen advanced medical VLMs. In contrast, our work identifies lesion-carrying patch representations and amplifies their pathological semantics at inference time, improving subtle lesion sensitivity without introducing a new detector or retraining the model.

\section{Methodology}

\begin{figure*}[t]
    \centering
    \includegraphics[width=\textwidth]{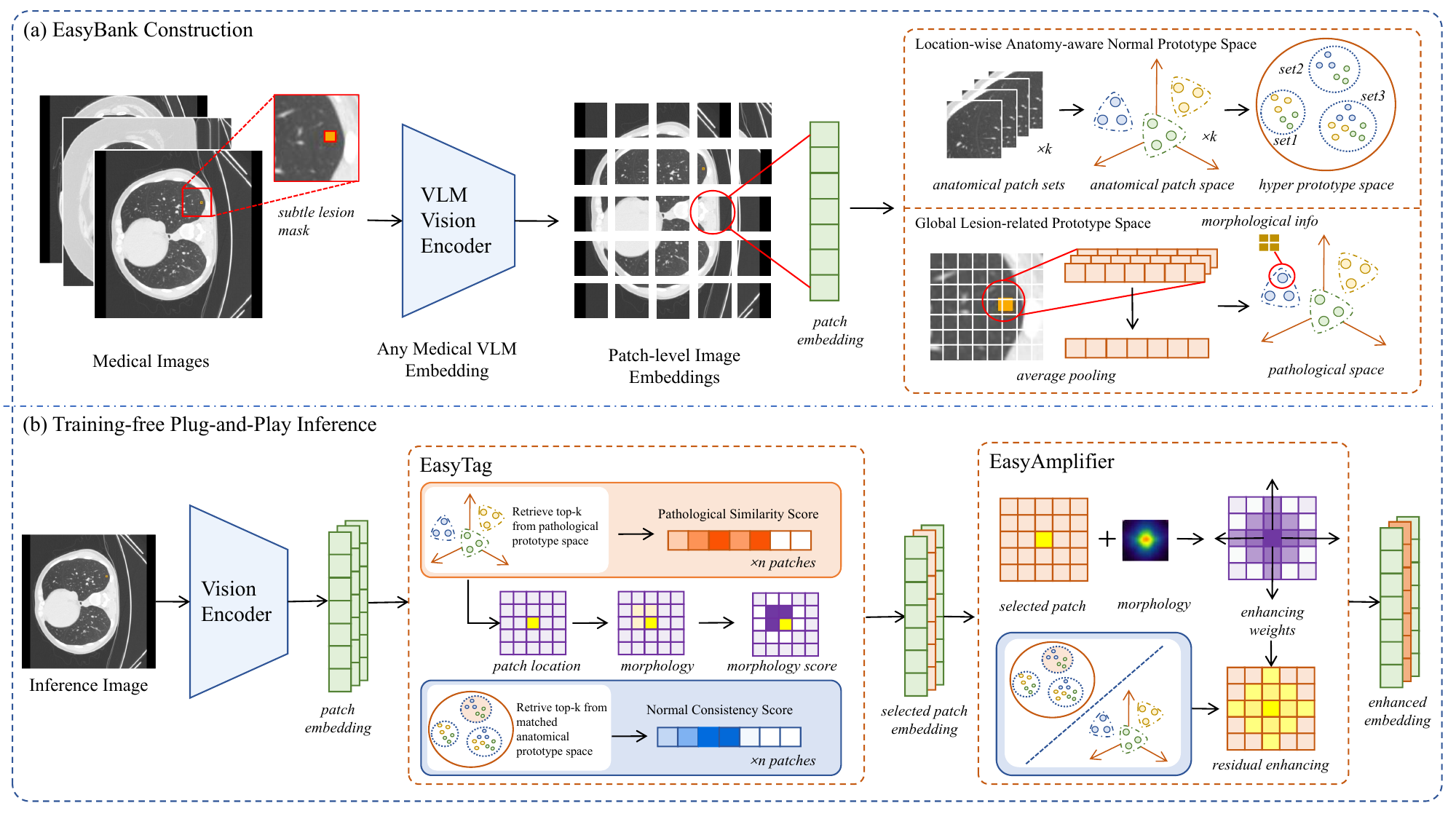}
    \vspace{-15pt}
    \caption{
    Overview of \textbf{EasyLens}.
    \textbf{(a) EasyBank Construction} builds an offline prototype space from CT images and lesion masks.
    \textbf{(b) Medical VLM Inference with EasyLens} selects lesion-relevant patches and amplifies their visual representations before feeding them into a frozen medical VLM for lesion-aware outputs.
    }
    \label{fig:flow}
    \vspace{-15pt}
\end{figure*}

\subsection{Overview}

As shown in Appendix Fig.~\ref{fig:representation_dilution}, our preliminary analysis suggests that subtle-lesion cues are not entirely absent from frozen medical VLMs. Instead, they can be partially preserved in patch-level visual representations but become underrepresented when local visual tokens are aggregated into global image representations. Since subtle lesions often occupy small regions, exhibit low contrast, or appear as weakly distinguishable cues within complex anatomical context, their representations can be easily mixed with normal anatomical patterns. This observation motivates us to strengthen lesion-relevant patch representations directly in the frozen visual embedding space, rather than updating the vision encoder through additional training.

To this end, we propose \textbf{EasyLens}, a training-free plug-and-play subtle-lesion representation amplifier for medical VLMs. As illustrated in Fig.~\ref{fig:flow}, EasyLens contains three components. First, \textbf{EasyBank} constructs a pathology-anatomy prototype space that stores lesion-related prototypes and anatomy-aware normal references. Second, \textbf{EasyTag} performs counterfactual prototype reasoning to select lesion-relevant patches by contrasting pathological similarity with location-matched normal consistency. Third, \textbf{EasyAmplifier} applies morphology-guided residual enhancement to the selected patch representations, increasing their contribution to the global image embedding while preserving the original visual context. The enhanced visual sequence is then passed to the subsequent components of the frozen medical VLM for downstream lesion-related tasks.

\subsection{EasyBank: Pathology-Anatomy Prototype Space}

EasyBank provides non-parametric visual references for distinguishing subtle lesion evidence from normal anatomical variation. Given a prototype construction set
$\mathcal{D}=\{(\mathbf{x}_i,\mathbf{m}_i)\}_{i=1}^{N}$,
where $\mathbf{x}_i$ denotes a medical image and $\mathbf{m}_i$ denotes its lesion mask, we extract patch-level visual representations from the frozen vision encoder:
\begin{equation}
    \mathbf{Z}_i
    =
    \mathcal{E}_{v}(\mathbf{x}_i)
    =
    [\mathbf{z}_{i}^{1},\mathbf{z}_{i}^{2},\ldots,\mathbf{z}_{i}^{P}]
    \in\mathbb{R}^{P\times d},
\end{equation}
where $P$ is the number of image patches and $d$ is the hidden dimension. The lesion masks are used only for constructing EasyBank and are not required during inference.

We project each lesion mask onto the patch grid and divide patches into lesion-related and normal anatomical sets:
\begin{equation}
    r_i^p
    =
    \frac{1}{|\Omega_p|}
    \sum_{\mathbf{u}\in\Omega_p}\mathbf{m}_i(\mathbf{u}),
    \quad
    \mathcal{P}_i
    =
    \{p\mid r_i^p>\tau\},
    \quad
    \mathcal{N}_i
    =
    \{1,\ldots,P\}\setminus\mathcal{P}_i,
\end{equation}
where $\Omega_p$ denotes the image region corresponding to the $p$-th patch, $r_i^p$ is the lesion occupancy ratio, and $\tau$ is the occupancy threshold.

EasyBank contains two complementary reference spaces. The first is a global lesion-related prototype space $\mathcal{C}^{L}$, which summarizes recurring pathological patterns. Since directly clustering all lesion patches would bias the prototype space toward large lesions, we first aggregate lesion-related patches within each lesion-containing image:
\begin{equation}
    \mathbf{h}_{i}^{L}
    =
    \frac{1}{|\mathcal{P}_i|}
    \sum_{p\in\mathcal{P}_i}\mathbf{z}_{i}^{p},
    \quad
    |\mathcal{P}_i|>0.
\end{equation}
The normalized image-level lesion representations are then clustered into $K_L$ lesion-related prototypes:
\begin{equation}
    \mathcal{C}^{L}
    =
    \operatorname{Cluster}
    \left(
    \{\hat{\mathbf{h}}_{i}^{L}\mid |\mathcal{P}_i|>0\},
    K_L
    \right).
\end{equation}
This design assigns equal weight to each lesion-containing image during prototype construction, preventing large lesions from dominating the prototype space simply because they occupy more patches.

The second reference space consists of location-wise anatomy-aware normal prototypes. Normal anatomical appearances vary substantially across spatial locations; for example, normal lung parenchyma, mediastinum, pleura, and abdominal organs may have very different visual representations. Therefore, using a single global normal prototype space would mix heterogeneous normal structures and provide ambiguous counterfactual references. EasyBank instead constructs a normal prototype subspace for each patch location:
\begin{equation}
    \mathcal{C}_{p}^{A}
    =
    \operatorname{Cluster}
    \left(
    \{\hat{\mathbf{z}}_{i}^{p}\mid p\in\mathcal{N}_i,\ i=1,\ldots,N\},
    K_A
    \right),
    \quad p=1,\ldots,P.
\end{equation}
Each $\mathcal{C}_{p}^{A}$ summarizes normal anatomical appearances at the same patch location and provides a location-matched reference for later counterfactual comparison.

In addition to prototype centers, EasyBank stores a lesion-support memory $\mathcal{L}_{k}$ and a morphology prior $\mathbf{M}_{k}$ for each lesion-related prototype. The support memory provides lesion-related reference embeddings for residual enhancement, while the morphology prior describes the spatial coherence of lesion patterns associated with the prototype. The final EasyBank is summarized as:
\begin{equation}
    \mathcal{B}
    =
    \left\{
    \mathcal{C}^{L},
    \{\mathcal{C}_{p}^{A}\}_{p=1}^{P},
    \{\mathcal{L}_{k},\mathbf{M}_{k}\}_{k=1}^{K_L}
    \right\}.
\end{equation}
Detailed clustering objectives, support-memory construction, and morphology-prior estimation are provided in Appendix Sec.~\ref{app:easybank}.

\subsection{EasyTag: Counterfactual Lesion-Relevant Patch Selection}

Given an inference image, EasyTag selects patches that are likely to contain subtle lesion evidence. For each patch representation $\mathbf{z}^{p}$, EasyTag compares it with two types of references in EasyBank: the global lesion-related prototypes $\mathcal{C}^{L}$ and the anatomy-aware normal prototypes $\mathcal{C}_{p}^{A}$ at the same patch location. The former measures whether the patch resembles pathological patterns, while the latter evaluates whether the patch can be explained by normal anatomy at the corresponding location.

Specifically, EasyTag retrieves the top-$M$ nearest lesion-related prototypes and top-$M$ nearest anatomy-aware normal prototypes for each patch, and computes the lesion similarity score $s_p^L$ and normal consistency score $s_p^A$. The counterfactual lesion relevance score is defined as:
\begin{equation}
    a_p
    =
    \sigma
    \left(
    \frac{s_p^L-s_p^A}{\tau_c}
    \right),
\end{equation}
where $\sigma(\cdot)$ is the sigmoid function and $\tau_c$ is a temperature parameter. A high score indicates that the patch is close to lesion-related prototypes but poorly explained by its location-matched normal references. In this sense, EasyTag implements counterfactual reasoning: it asks whether a suspicious patch still appears abnormal after being compared with normal anatomical appearances from the same location.

Patch-wise scores may be noisy when lesion evidence is weak. Moreover, subtle lesions often appear as spatially coherent local structures rather than isolated patches. To include weak but morphology-consistent lesion cues, EasyTag calibrates the initial scores using morphology priors stored in EasyBank. We first select a high-confidence seed set $\mathcal{S}_0$ based on $\{a_p\}_{p=1}^{P}$. For each seed patch $p$, we identify its nearest lesion-related prototype:
\begin{equation}
    k^{*}(p)
    =
    \arg\max_{k}
    \operatorname{sim}(\mathbf{z}^{p},\mathbf{c}_{k}^{L}).
\end{equation}
The morphology prior associated with this prototype is then used to propagate confidence from seed patches to spatially coherent neighboring patches:
\begin{equation}
    \tilde{a}_{q}
    =
    a_q
    +
    \lambda
    \max_{p\in\mathcal{S}_0}
    \left[
    a_p\mathbf{M}_{k^{*}(p)}(q-p)
    \right],
    \quad
    \mathcal{S}^{C}
    =
    \operatorname{TopK}_{q}
    \tilde{a}_{q},
\end{equation}
where $\lambda$ controls the calibration strength, $\mathbf{M}_{k^{*}(p)}(q-p)$ denotes the morphology-prior value at the relative offset from seed patch $p$ to patch $q$, and $\mathcal{S}^{C}$ is the final candidate set. This calibration allows EasyTag to select sparse but spatially coherent lesion evidence while avoiding blind amplification of normal tissues. The detailed retrieval procedure and score computation are provided in Appendix Sec.~\ref{app:easytag}.

\subsection{EasyAmplifier: Morphology-Guided Residual Enhancement}

After EasyTag selects lesion-relevant candidate patches, EasyAmplifier strengthens their representations before they are passed to downstream VLM components. The goal is not to replace the original visual embeddings, but to inject lesion-related residual directions into patches that are supported by both counterfactual evidence and morphology priors. This design preserves the anatomical context encoded by the frozen vision encoder while increasing the contribution of subtle lesion cues to the global image embedding.

For each selected candidate patch $p\in\mathcal{S}^{C}$, EasyAmplifier retrieves a lesion-related reference from the support memory $\mathcal{L}_{k^{*}(p)}$ associated with its recalled lesion prototype. This reference provides a prototype-consistent pathological direction in the frozen embedding space. To extend enhancement beyond isolated high-confidence patches, EasyAmplifier computes, for every patch $q$, the strongest morphology-consistent support from the selected candidates:
\begin{equation}
    p^{*}(q)
    =
    \arg\max_{p\in\mathcal{S}^{C}}
    \tilde{a}_{p}\mathbf{M}_{k^{*}(p)}(q-p),
    \quad
    w_q
    =
    \max_{p\in\mathcal{S}^{C}}
    \tilde{a}_{p}\mathbf{M}_{k^{*}(p)}(q-p).
\end{equation}
Here, $p^{*}(q)$ identifies the selected candidate that provides the strongest morphology-supported evidence for patch $q$, and $w_q$ measures the strength of this support. If $w_q$ is sufficiently large, patch $q$ is considered part of a morphology-consistent lesion region.

EasyAmplifier then retrieves a lesion-related reference embedding $\mathbf{r}_q$ from the support memory of the strongest recalled prototype and updates the patch representation through a score-weighted residual enhancement:
\begin{equation}
    \bar{\mathbf{z}}^{q}
    =
    \begin{cases}
    \mathbf{z}^{q}
    +
    \alpha w_q(\mathbf{r}_q-\mathbf{z}^{q}),
    &
    w_q>\eta,\\
    \mathbf{z}^{q},
    &
    w_q\leq\eta,
    \end{cases}
\end{equation}
where $\alpha$ controls the amplification strength and $\eta$ prevents low-confidence patches from being modified. The residual direction $(\mathbf{r}_q-\mathbf{z}^{q})$ moves the patch toward a lesion-related reference while retaining its original embedding as the base representation. Thus, high-confidence and morphology-supported lesion cues are enhanced more strongly, whereas unrelated anatomical regions remain unchanged.

Finally, EasyAmplifier outputs the enhanced visual sequence
$\bar{\mathbf{Z}}=[\bar{\mathbf{z}}^{1},\ldots,\bar{\mathbf{z}}^{P}]$,
which replaces the original patch sequence before being passed to the subsequent medical VLM components. Since EasyLens relies only on prototype retrieval, counterfactual scoring, morphology-guided weighting, and residual enhancement, it requires no gradient-based optimization, no inference-time lesion annotations, and no modification of the frozen VLM.

\begin{table*}[t]
  \centering
  \small
  \setlength{\tabcolsep}{2.8pt}
  \renewcommand{\arraystretch}{1.16}
    \begin{tabularx}{\textwidth}{
        @{\hspace{5pt}}
        l|
        *{3}{>{\centering\arraybackslash}X}|
        *{3}{>{\centering\arraybackslash}X}|
        *{3}{>{\centering\arraybackslash}X}
        @{}
    }
    \hline\hline
    \multirow{2}{*}{\textbf{Models}} &
    \multicolumn{3}{c|}{\textbf{ReX}} &
    \multicolumn{3}{c|}{\textbf{LIDC}} &
    \multicolumn{3}{c}{\textbf{Abdomen}} \\
    \cline{2-10}
    & \textbf{Stat.} & \textbf{Sel.} & \textbf{Gen.}
    & \textbf{Stat.} & \textbf{Sel.} & \textbf{Gen.}
    & \textbf{Stat.} & \textbf{Sel.} & \textbf{Gen.} \\
    \hline

    LLaVA-Med     & 0.00 & 1.11 & 3.93 & 0.00 & 31.58 & 33.92 & 1.16 & 14.42 & 36.60 \\
    RadFM         & 46.96 & 7.78 & 2.96 & 0.00 & 21.05 & 13.75 & 10.47 & 11.54 & 9.60 \\
    Lingshu       & 7.48 & 25.56 & 0.33 & 20.45 & 33.83 & 16.20 & 37.21 & 53.85 & 25.04 \\
    Hulu-Med      & 64.63 & 27.78 & 0.88 & 28.79 & 30.83 & 23.87 & 51.16 & 52.88 & 39.89 \\
    MedGemma      & 40.82 & 24.44 & 4.55 & 25.00 & 21.05 & 35.94 & 27.91 & 45.19 & 26.11 \\
    MedGemma1.5   & 42.86 & 23.33 & 4.41 & 20.45 & 27.82 & 41.93 & 15.12 & 49.04 & 38.18 \\
    \hline

    MedKLIP       & 0.00 & 17.78 & - & 0.00 & 27.07 & - & 0.00 & 28.85 & - \\
    KAD           & 0.00 & 13.33 & - & 0.00 & 16.54 & - & 0.00 & 5.77 & - \\
    \hline

    \textbf{EasyLens} & \textbf{66.67} & \textbf{31.11} & \textbf{5.15}
                      & \textbf{30.30} & \textbf{36.09} & \textbf{45.86}
                      & \textbf{52.33} & \textbf{55.77} & \textbf{40.67} \\
    \hline\hline
  \end{tabularx}

  \vspace{6pt}
  \caption{Comparison of different medical VLMs on ReX, LIDC, and Abdomen datasets.}
  \label{tab:main_comparison}
\end{table*}

\begin{table*}[t]
  \centering
  \tiny
  \setlength{\tabcolsep}{5.0pt}
  \renewcommand{\arraystretch}{1.18}

  \resizebox{\textwidth}{!}{
  \begin{tabular}{
    @{\hspace{4pt}}
    l|c|
    cccc|cccc
    @{\hspace{4pt}}
  }
    \hline\hline
    \multirow{2}{*}{\textbf{Models}} &
    \multirow{2}{*}{\textbf{w/ EasyLens}} &
    \multicolumn{4}{c|}{\textbf{ReX}} &
    \multicolumn{4}{c}{\textbf{Kvasir-SEG}} \\
    \cline{3-10}
    & & Stat. & Sel. & Gen. & Inf.(s)
      & BL-1 & MTR & RG-L & Inf.(s) \\
    \hline

    \multirow{2}{*}{LLaVA-Med}
    & \ding{55} & 0.00 & 1.11 & 3.93  & 1.03 & 11.49 & 22.69 & 8.61  & 3.33 \\
    \cline{2-10}
    & \ding{51} & 0.00 & 1.11 & \textbf{32.37} & \textbf{1.13} & \textbf{12.22} & \textbf{24.35} & 8.56  & 2.80 \\
    \hline

    \multirow{2}{*}{RadFM}
    & \ding{55} & 46.94 & 7.78  & 2.96 & 2.68 & 1.60 & 1.31 & 2.41 & 0.57 \\
    \cline{2-10}
    & \ding{51} & \textbf{48.98} & \textbf{10.00} & \textbf{4.12} & \textbf{2.75} & \textbf{2.28} & \textbf{1.44} & \textbf{2.61} & 0.49 \\
    \hline

    \multirow{2}{*}{Lingshu}
    & \ding{55} & 7.48 & 25.56 & 0.33 & 0.46 & 20.06 & 10.81 & 20.38 & 0.96 \\
    \cline{2-10}
    & \ding{51} & 7.48 & \textbf{27.78} & \textbf{8.51} & \textbf{0.49} & \textbf{26.69} & \textbf{13.89} & \textbf{30.00} & 0.63 \\
    \hline

    \multirow{2}{*}{MedGemma}
    & \ding{55} & 40.82 & 24.44 & 4.55 & 0.78 & 13.99 & 25.38 & 15.59 & 1.43 \\
    \cline{2-10}
    & \ding{51} & \textbf{46.26} & \textbf{25.56} & \textbf{6.73} & \textbf{1.48} & 13.81 & \textbf{26.00} & \textbf{16.19} & \textbf{1.81} \\
    \hline

    \multirow{2}{*}{MedGemma1.5}
    & \ding{55} & 42.86 & 23.33 & 4.41 & 1.80 & 9.56  & 15.87 & 15.39 & 1.87 \\
    \cline{2-10}
    & \ding{51} & \textbf{66.67} & \textbf{31.11} & \textbf{5.15} & \textbf{2.38} & \textbf{10.30} & \textbf{16.55} & \textbf{16.18} & \textbf{2.01} \\
    \hline\hline
  \end{tabular}
  }
  \caption{Comparison of different medical VLMs with and without the proposed module on ReX and Kvasir-SEG datasets.}
  \label{tab:comparison_wo_w}
\end{table*}

\section{Experiments}

\subsection{Datasets and Evaluation Tasks}

We evaluate EasyLens on two dataset groups spanning the lesion spectrum described in the introduction. The first group focuses on \emph{subtle lesions}, for which we construct a unified benchmark from ReXGroundingCT, LIDC-IDRI, and AbdomenAtlas 3.0 Mini, denoted as ReX, LIDC, and Abdomen, respectively. This benchmark contains three task types: regional statistics (Stat.), region selection (Sel.), and lesion-aware report generation (Gen.). The second group contains more salient or regular lesion datasets, including MIMIC-CXR, Kvasir-SEG, and BKAI-Polyp, where we evaluate standard medical report generation using BLEU-1, BLEU-4, METEOR, and ROUGE-L. These datasets verify that EasyLens does not over-specialize to subtle abnormalities and preserves general report-generation ability when lesions are visually more apparent. Detailed dataset construction, task definitions, evaluation metrics, and benchmark distributions are provided in Appendix Sec.~\ref{app:dataset_details}.

\subsection{Implementation and Experimental Setup}

Unless otherwise specified, EasyLens denotes MedGemma1.5 equipped with the proposed frozen-backbone inference-time adapter in Table~\ref{tab:main_comparison}, while Table~\ref{tab:comparison_wo_w} applies the same adapter to each medical VLM backbone. EasyLens keeps the visual encoder, multimodal projector, and language decoder fixed, and only calibrates four interface parameters, i.e., residual strength $\beta$, selected-token budget $k$, local seed budget $s$, and support retrieval budget $m$, as detailed in Appendix~\ref{app:hyperparam}. We report subtle-lesion results in Table~\ref{tab:main_comparison}, backbone transfer in Table~\ref{tab:comparison_wo_w}, general and salient-lesion report generation in Appendix~\ref{app:general_large}, and interface ablations in Appendix~\ref{app:ablation_detail}. For report-generation tasks, we use raw benchmark prompts, a 160-token generation limit, and the evaluation-only lesion-aware probe described in Appendix~\ref{app:report_probe}.

\subsection{Quantitative Evaluation}
\label{sec:quant_eval}

\noindent\textbf{EasyLens yields consistent improvements across all subtle-lesion evaluation settings.}
Table~\ref{tab:main_comparison} reports results on ReX, LIDC, and Abdomen, covering lesion status recognition, lesion-aware region selection, and report generation. EasyLens achieves the best performance in all nine dataset--task combinations. Relative to the strongest competing baseline in each setting, EasyLens improves ReX by 2.04, 3.33, and 0.60 points on Stat., Sel., and Gen., respectively. The corresponding gains are 1.51, 2.26, and 3.93 points on LIDC, and 1.17, 1.92, and 0.78 points on Abdomen. These results indicate that EasyLens improves the recognition of small, low-contrast, and spatially sparse lesions that are difficult for existing medical VLMs to capture reliably.

\noindent\textbf{EasyLens transfers effectively across frozen medical VLM backbones.}
Table~\ref{tab:comparison_wo_w} further evaluates EasyLens when attached to different frozen backbones, including LLaVA-Med, RadFM, Lingshu, MedGemma, and MedGemma1.5. On ReX, EasyLens improves report-generation performance for all five backbones, with particularly large gains for LLaVA-Med and Lingshu, increasing Gen. from 3.93 to 32.37 and from 0.33 to 8.51, respectively. For the strongest backbone, MedGemma1.5, EasyLens improves Stat./Sel./Gen. from 42.86/23.33/4.41 to 66.67/31.11/5.15. These results show that the proposed module is not tied to a specific backbone architecture.

\noindent\textbf{EasyLens preserves general report-generation capability beyond subtle-lesion tasks.}
As shown in Appendix~\ref{app:general_large}, EasyLens remains comparable to MedGemma1.5 on MIMIC-CXR and improves performance on Kvasir-SEG and BKAI-Polyp across all non-zero lexical metrics. Together with the backbone-transfer results in Table~\ref{tab:comparison_wo_w}, these findings suggest that amplifying lesion-relevant visual evidence does not substantially disrupt the original reporting behavior of the frozen VLM, while providing consistent benefits for subtle-lesion perception.

\begin{table}[t]
  \centering
  \small
  \setlength{\tabcolsep}{3.5pt}
  \renewcommand{\arraystretch}{1.14}

  \newcolumntype{Y}{>{\centering\arraybackslash}X}
  \newcolumntype{L}[1]{>{\raggedright\arraybackslash}p{#1}}

  \begin{tabularx}{\linewidth}{
    @{}
    L{0.18\linewidth}YYY|
    L{0.18\linewidth}YYY
    @{}
  }
    \hline\hline
    \multicolumn{4}{c|}{\textbf{EasyTag Selector Ablation}} &
    \multicolumn{4}{c}{\textbf{EasyAmplifier Amplifier Ablation}} \\
    \hline

    \multirow{2}{*}{\textbf{Models}} &
    \multicolumn{3}{c|}{\textbf{ReX}} &
    \multirow{2}{*}{\textbf{Models}} &
    \multicolumn{3}{c}{\textbf{ReX}} \\
    \cline{2-4}\cline{6-8}
    & \textbf{Stat.} & \textbf{Sel.} & \textbf{Gen.}
    & & \textbf{Stat.} & \textbf{Sel.} & \textbf{Gen.} \\
    \hline

    MedGemma1.5
    & 42.86 & 23.33 & 4.41
    & MedGemma1.5
    & 42.86 & 23.33 & 4.41 \\

    \hspace*{1.0em}w/ MLP Selector
    & 69.39 & 27.78 & 4.13
    & \hspace*{1.0em}w/o Residual
    & 64.63 & \textbf{33.33} & 4.89 \\

    \hspace*{1.0em}w/ GT Masks
    & \textbf{71.43} & \textbf{32.22} & \textbf{5.36}
    & \hspace*{1.0em}w/o Morphology
    & 57.14 & 28.89 & 3.17 \\

    \hspace*{1.0em}w/ EasyTag
    & 66.67 & 31.33 & 5.15
    & \hspace*{1.0em}w/ EasyAmplifier
    & \textbf{66.67} & 31.11 & \textbf{5.15} \\
    \hline\hline
  \end{tabularx}
  \caption{Ablation studies of the EasyTag and EasyAmplifier on the ReX dataset.}
  \label{tab:cprs_EasyAmplifier_ablation}
  \vspace{-10pt}
\end{table}

\subsection{Ablation Study}
\label{sec:ablation}

\noindent \textbf{EasyTag provides training-free selection competitive with supervised alternatives.}
As shown in Table~\ref{tab:cprs_EasyAmplifier_ablation}, EasyTag improves MedGemma1.5 on ReX, increasing Stat. from 42.86 to 66.67, Sel. from 23.33 to 31.33, and Gen. from 4.41 to 5.15. The MLP Selector is a trainable patch-scoring module supervised to identify lesion-relevant tokens. Although it obtains a slightly higher Stat. score of 69.39, its Sel. and Gen. scores decrease to 27.78 and 4.13. This comparison shows that EasyTag achieves competitive performance without training, while providing stronger lesion-aware evidence for selection and generation. The GT-mask setting selects lesion-overlapping tokens using ground-truth masks and serves only as a mask-guided upper-reference. Its results, 71.43 on Stat., 32.22 on Sel., and 5.36 on Gen., are close to EasyTag, further indicating that EasyTag recovers most useful lesion evidence without inference-time masks.

\noindent \textbf{EasyAmplifier benefits from residual preservation and morphology-aware enhancement.}
Table~\ref{tab:cprs_EasyAmplifier_ablation} also evaluates the amplification design. Removing the residual formulation reduces Stat. from 66.67 to 64.63 and Gen. from 5.15 to 4.89, while Sel. increases from 31.11 to 33.33. Removing morphology modeling causes a broader drop, reducing Stat. to 57.14, Sel. to 28.89, and Gen. to 3.17. These results suggest that morphology-aware enhancement is the main source of subtle-lesion amplification, while the residual path helps preserve the original visual semantics. Additional interface-level ablations are provided in Appendix~\ref{app:ablation_detail}.

\begin{figure*}[t]
    \centering
    \includegraphics[width=\textwidth]{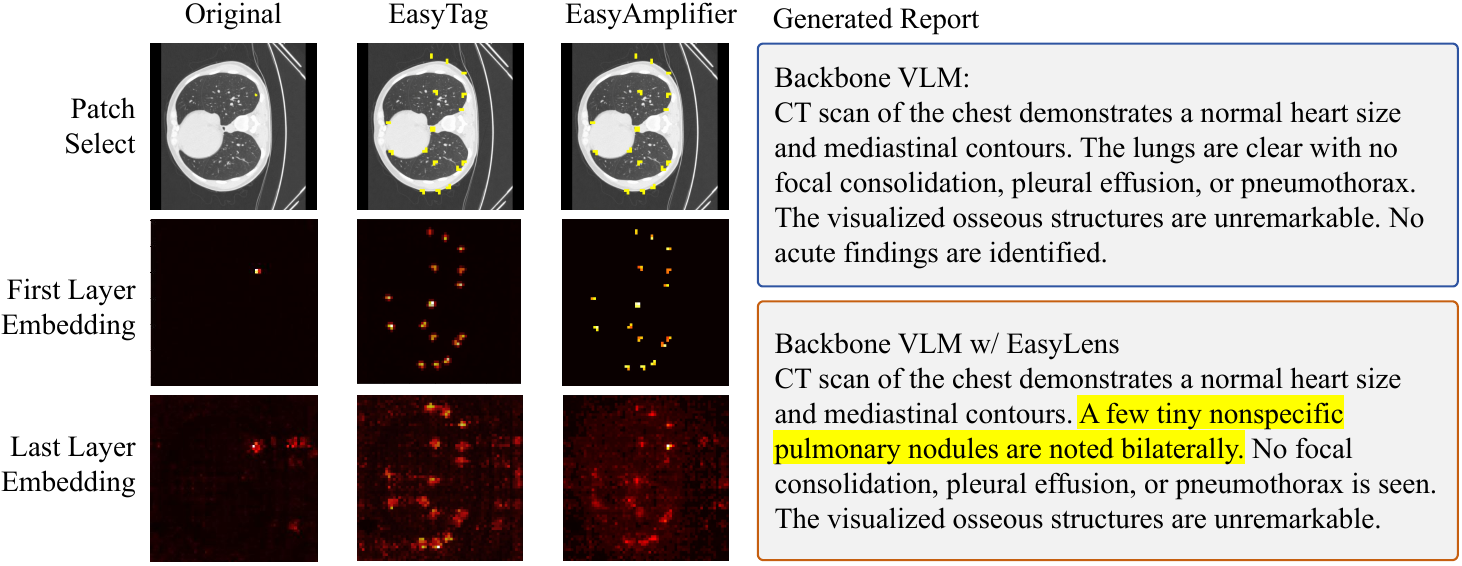}
    \vspace{-15pt}
    \caption{Case study of EasyLens on subtle-lesion perception.}
    \label{fig:cs1}
    \vspace{-10pt}
\end{figure*}

\subsection{Case Study}
\label{sec:case_study}

\noindent \textbf{Qualitative evidence for lesion-relevant token selection and amplification.}
Figure~\ref{fig:cs1} provides a qualitative example of a CT case containing a subcentimeter pulmonary nodule. The selected tokens produced by EasyTag overlap with the suspected nodule region, supporting the claim that the training-free selector can identify micro-lesion-related visual evidence from frozen representations. After EasyAmplifier, the response associated with the selected lesion patches becomes stronger and more spatially concentrated on the nodule region, indicating that morphology-guided residual enhancement increases the lesion-semantic contribution of the relevant patches. This visual change is consistent with the generated report, where the EasyLens-enhanced model mentions the tiny pulmonary nodule, while the baseline model omits this finding.

\section{Conclusion}

We presented \textbf{EasyLens}, a training-free plug-and-play amplifier that improves subtle-lesion perception in frozen medical VLMs. EasyLens builds a pathology-anatomy prototype space, selects lesion-relevant tokens through counterfactual prototype reasoning, and strengthens them with morphology-guided residual enhancement. Experiments across subtle-lesion benchmarks and multiple VLM backbones show consistent gains in lesion status recognition, region selection, and lesion-aware report generation without updating model parameters. These results show that \textbf{EasyLens} can expose and amplify weak lesion evidence already encoded in frozen visual tokens, enabling medical VLMs to better recognize and report clinically important micro-lesions without model-specific retraining.

\clearpage
\bibliographystyle{plainnat}
\bibliography{references}
\clearpage
\appendix
\section{Additional Method Details}
\label{app:method_details}

\subsection{Preliminary Analysis of Representation Dilution}
\label{app:representation_dilution}

\begin{figure}[H]
    \centering
    \begin{minipage}{0.58\linewidth}
        \centering
        \includegraphics[width=\linewidth]{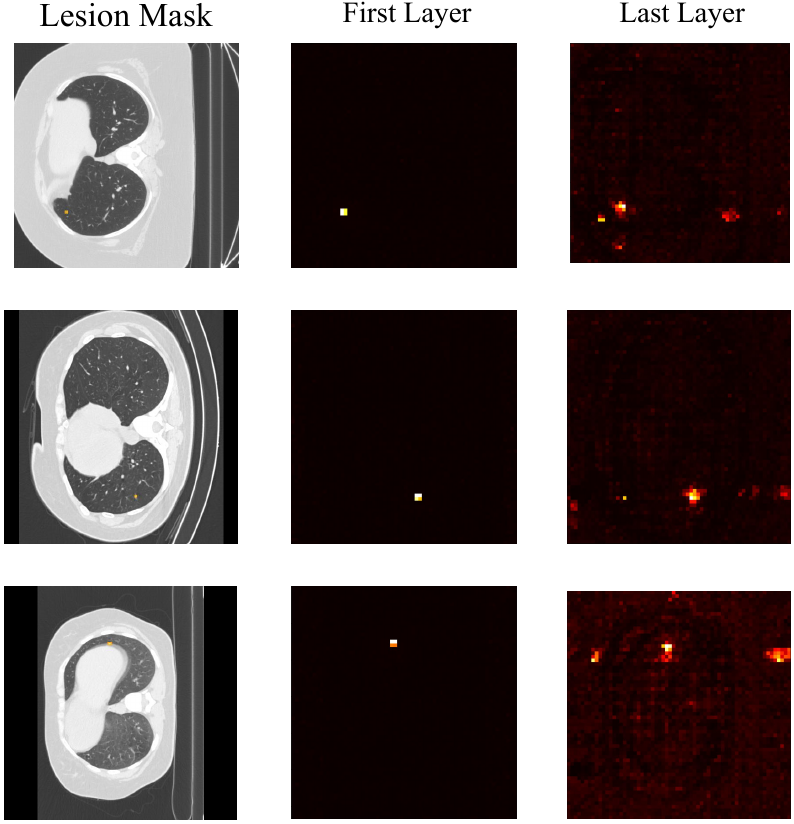}
        \caption{Subtle-lesion representation dilution in medical VLMs.}
        \label{fig:representation_dilution}
    \end{minipage}
\end{figure}

Appendix Fig.~\ref{fig:representation_dilution} provides the empirical motivation for EasyLens. The analysis shows that subtle-lesion cues are not completely absent in frozen medical VLM representations. Instead, their patch-level evidence can be weakly encoded but becomes less separable from normal anatomical regions during visual-token aggregation. This observation motivates a training-free representation enhancement strategy that operates directly on patch-level visual embeddings.

\subsection{Detailed Construction of EasyBank}
\label{app:easybank}

EasyBank is a non-parametric pathology-anatomy prototype space constructed from patch-level representations extracted by a frozen medical VLM. Given a prototype construction set
$\mathcal{D}=\{(\mathbf{x}_i,\mathbf{m}_i)\}_{i=1}^{N}$,
where $\mathbf{x}_i$ is a medical image and $\mathbf{m}_i\in\{0,1\}^{H\times W}$ is its lesion mask, we extract visual patch representations:
\begin{equation}
    \mathbf{Z}_i
    =
    \mathcal{E}_{v}(\mathbf{x}_i)
    =
    [\mathbf{z}_{i}^{1},\mathbf{z}_{i}^{2},\ldots,\mathbf{z}_{i}^{P}]
    \in\mathbb{R}^{P\times d}.
\end{equation}

\paragraph{Patch-level lesion assignment.}
Let $\Omega_p$ denote the image region covered by the $p$-th visual patch. We project the lesion mask onto the patch grid by computing the lesion occupancy ratio:
\begin{equation}
    r_i^p
    =
    \frac{1}{|\Omega_p|}
    \sum_{\mathbf{u}\in\Omega_p}
    \mathbf{m}_i(\mathbf{u}).
\end{equation}
Patches are divided into lesion-related and normal anatomical regions:
\begin{equation}
    \mathcal{P}_i
    =
    \{p\mid r_i^p>\tau\},
    \quad
    \mathcal{N}_i
    =
    \{1,\ldots,P\}\setminus\mathcal{P}_i,
\end{equation}
where $\tau$ is the lesion occupancy threshold.

\paragraph{Lesion-related prototype construction.}
Directly clustering all lesion patches would bias prototype construction toward salient lesions, since they contribute more patch embeddings. To reduce this imbalance, we represent each lesion-containing image using the mean embedding of its lesion-related patches:
\begin{equation}
    \mathbf{h}_{i}^{L}
    =
    \frac{1}{|\mathcal{P}_i|}
    \sum_{p\in\mathcal{P}_i}
    \mathbf{z}_{i}^{p},
    \quad
    |\mathcal{P}_i|>0.
\end{equation}
The normalized lesion representations are collected into:
\begin{equation}
    \mathcal{B}^{L}
    =
    \{
    \hat{\mathbf{h}}_{i}^{L}
    \mid
    |\mathcal{P}_i|>0,\ i=1,\ldots,N
    \}.
\end{equation}
We cluster this buffer into $K_L$ lesion-related prototypes:
\begin{equation}
    \mathcal{C}^{L}
    =
    \{
    \mathbf{c}_{1}^{L},
    \mathbf{c}_{2}^{L},
    \ldots,
    \mathbf{c}_{K_L}^{L}
    \}.
\end{equation}
The clustering objective is:
\begin{equation}
    \min_{\{\mathbf{c}_{k}^{L}\}_{k=1}^{K_L}}
    \sum_{\mathbf{h}\in\mathcal{B}^{L}}
    \min_{k}
    \left\|
    \mathbf{h}
    -
    \hat{\mathbf{c}}_{k}^{L}
    \right\|_{2}^{2}.
\end{equation}
Each lesion-related prototype summarizes a recurring pathological representation pattern discovered from frozen VLM embeddings.

\paragraph{Lesion-support memory.}
For each lesion-related prototype, we retain a support memory containing patch-level lesion embeddings from construction samples assigned to that prototype. Let
\begin{equation}
    \pi(i)
    =
    \arg\min_{k}
    \left\|
    \hat{\mathbf{h}}_{i}^{L}
    -
    \hat{\mathbf{c}}_{k}^{L}
    \right\|_{2}^{2}
\end{equation}
denote the prototype assignment of image $i$. The support memory for prototype $k$ is:
\begin{equation}
    \mathcal{L}_{k}
    =
    \left\{
    \hat{\mathbf{z}}_{i}^{p}
    \ \middle|\
    \pi(i)=k,\ p\in\mathcal{P}_i
    \right\}.
\end{equation}
This memory provides lesion-related reference embeddings for EasyAmplifier without adding trainable parameters.

\paragraph{Morphology prior estimation.}
Subtle lesions are often spatially coherent rather than isolated at a single patch. For each lesion-related prototype, we estimate a morphology prior from the patch-level lesion supports of its assigned construction samples. For a relative patch offset $\delta$, the prior is computed as:
\begin{equation}
    \mathbf{M}_{k}(\delta)
    =
    \frac{1}{Z_k}
    \sum_{\pi(i)=k}
    \sum_{p\in\mathcal{P}_i}
    \mathbf{1}[p+\delta\in\mathcal{P}_i],
\end{equation}
where $\mathbf{1}[\cdot]$ is the indicator function and $Z_k$ is a normalization factor. We normalize $\mathbf{M}_{k}$ to $[0,1]$. This prior estimates how likely neighboring patches are to belong to the same lesion pattern when a patch recalls prototype $k$.

\paragraph{Anatomy-aware normal prototype construction.}
Normal anatomical appearances vary substantially across spatial locations. Therefore, EasyBank constructs location-wise normal prototype subspaces rather than using a single global normal prototype set. For each patch location $p$, we collect normal embeddings from the same position:
\begin{equation}
    \mathcal{B}_{p}^{A}
    =
    \{
    \hat{\mathbf{z}}_{i}^{p}
    \mid
    p\in\mathcal{N}_i,\ i=1,\ldots,N
    \}.
\end{equation}
Each location-specific buffer is clustered into $K_A$ anatomy-aware normal prototypes:
\begin{equation}
    \mathcal{C}_{p}^{A}
    =
    \{
    \mathbf{c}_{p,1}^{A},
    \mathbf{c}_{p,2}^{A},
    \ldots,
    \mathbf{c}_{p,K_A}^{A}
    \}.
\end{equation}
The clustering objective is:
\begin{equation}
    \min_{\{\mathbf{c}_{p,k}^{A}\}_{k=1}^{K_A}}
    \sum_{\mathbf{z}\in\mathcal{B}_{p}^{A}}
    \min_{k}
    \left\|
    \mathbf{z}
    -
    \hat{\mathbf{c}}_{p,k}^{A}
    \right\|_{2}^{2}.
\end{equation}

The final EasyBank is:
\begin{equation}
    \mathcal{B}
    =
    \left\{
    \mathcal{C}^{L},
    \{\mathcal{C}_{p}^{A}\}_{p=1}^{P},
    \{\mathcal{L}_{k},\mathbf{M}_{k}\}_{k=1}^{K_L}
    \right\}.
\end{equation}
It provides lesion-related prototypes, anatomy-aware normal references, lesion-support memories, and morphology priors for EasyTag and EasyAmplifier.

\subsection{Detailed Counterfactual Prototype Reasoning in EasyTag}
\label{app:easytag}

Given an inference image, the frozen vision encoder produces patch-level representations:
\begin{equation}
    \mathbf{Z}
    =
    [\mathbf{z}^{1},\mathbf{z}^{2},\ldots,\mathbf{z}^{P}].
\end{equation}
EasyTag evaluates each patch from two complementary perspectives. A lesion-relevant patch should be close to lesion-related prototypes in $\mathcal{C}^{L}$, while being poorly explained by anatomy-aware normal prototypes $\mathcal{C}_{p}^{A}$ from the same spatial location.

For patch $\mathbf{z}^{p}$, EasyTag retrieves its top-$M$ nearest lesion-related prototypes and top-$M$ nearest anatomy-aware normal prototypes:
\begin{equation}
    \mathcal{R}_{p}^{L}
    =
    \operatorname{TopM}(\mathcal{C}^{L},\mathbf{z}^{p}),
    \quad
    \mathcal{R}_{p}^{A}
    =
    \operatorname{TopM}(\mathcal{C}_{p}^{A},\mathbf{z}^{p}).
\end{equation}
The lesion similarity score and normal consistency score are:
\begin{equation}
    s_{p}^{L}
    =
    \frac{1}{M}
    \sum_{\mathbf{c}\in\mathcal{R}_{p}^{L}}
    \operatorname{sim}(\mathbf{z}^{p},\mathbf{c}),
    \quad
    s_{p}^{A}
    =
    \frac{1}{M}
    \sum_{\mathbf{c}\in\mathcal{R}_{p}^{A}}
    \operatorname{sim}(\mathbf{z}^{p},\mathbf{c}).
\end{equation}
The initial counterfactual lesion relevance score is:
\begin{equation}
    a_p
    =
    \sigma
    \left(
    \frac{s_p^L-s_p^A}{\tau_c}
    \right).
\end{equation}
A normal patch is expected to have high normal consistency with $\mathcal{C}_{p}^{A}$, whereas a lesion-relevant patch should exhibit higher pathological similarity and lower consistency with its normal anatomical reference.

\paragraph{Morphology-guided score calibration.}
Patch-wise counterfactual scores may be noisy when lesion evidence is weak. To include spatially coherent but low-confidence lesion evidence, EasyTag calibrates scores using morphology priors. We first select high-confidence seed patches:
\begin{equation}
    \mathcal{S}_{0}
    =
    \operatorname{TopK}_{0}
    \{a_p\}_{p=1}^{P}.
\end{equation}
For each seed patch $p$, its nearest lesion-related prototype is:
\begin{equation}
    k^{*}(p)
    =
    \arg\max_{k}
    \operatorname{sim}(\mathbf{z}^{p},\mathbf{c}_{k}^{L}).
\end{equation}
The calibrated score for patch $q$ is:
\begin{equation}
    \tilde{a}_{q}
    =
    a_q
    +
    \lambda
    \max_{p\in\mathcal{S}_{0}}
    \left[
    a_p\mathbf{M}_{k^{*}(p)}(q-p)
    \right],
\end{equation}
where $\lambda$ controls the strength of morphology-guided calibration. The final candidate set is:
\begin{equation}
    \mathcal{S}^{C}
    =
    \operatorname{TopK}_{q\in\{1,\ldots,P\}}
    \tilde{a}_{q}.
\end{equation}
This procedure encourages EasyTag to select spatially coherent lesion regions rather than isolated high-scoring patches.

\subsection{Detailed Morphology-Guided Residual Enhancement in EasyAmplifier}
\label{app:easyamplifier}

EasyAmplifier strengthens lesion-relevant representations selected by EasyTag. Although the selected patches may already contain lesion evidence, their original embeddings can still be weak due to representation dilution during visual-token aggregation. EasyAmplifier therefore enhances them through residual updates in the visual embedding space, rather than replacing the original features.

\paragraph{Candidate-level residual direction.}
For each selected candidate patch $p\in\mathcal{S}^{C}$, EasyTag provides its calibrated score $\tilde{a}_p$ and recalled lesion prototype $k^{*}(p)$. EasyAmplifier retrieves the closest lesion-related reference from the corresponding support memory:
\begin{equation}
    \mathbf{r}_{p}
    =
    \arg\max_{\mathbf{u}\in\mathcal{L}_{k^{*}(p)}}
    \operatorname{sim}(\mathbf{z}^{p},\mathbf{u}).
\end{equation}
The residual direction is:
\begin{equation}
    \Delta_p
    =
    \mathbf{r}_{p}-\mathbf{z}^{p}.
\end{equation}
The candidate patch can be enhanced as:
\begin{equation}
    \mathbf{z}^{p,+}
    =
    \mathbf{z}^{p}
    +
    \alpha\tilde{a}_{p}\Delta_p,
    \quad
    p\in\mathcal{S}^{C},
\end{equation}
where $\alpha$ controls the residual amplification strength. This update moves the candidate patch toward a lesion-related reference while preserving its original anatomical context.

\paragraph{Morphology-guided propagation.}
Enhancing only selected candidates may miss nearby patches that contain weaker but spatially consistent lesion evidence. EasyAmplifier therefore propagates residual enhancement according to the morphology prior recalled by each selected candidate. For every patch $q$, we compute:
\begin{equation}
    p^{*}(q)
    =
    \arg\max_{p\in\mathcal{S}^{C}}
    \tilde{a}_{p}\mathbf{M}_{k^{*}(p)}(q-p),
    \quad
    w_q
    =
    \max_{p\in\mathcal{S}^{C}}
    \tilde{a}_{p}\mathbf{M}_{k^{*}(p)}(q-p).
\end{equation}
When $w_q>\eta$, patch $q$ is treated as morphologically supported by the selected lesion evidence. Its lesion-related reference is retrieved from the support memory of the strongest recalled prototype:
\begin{equation}
    \mathbf{r}_q
    =
    \arg\max_{\mathbf{u}\in\mathcal{L}_{k^{*}(p^{*}(q))}}
    \operatorname{sim}(\mathbf{z}^{q},\mathbf{u}).
\end{equation}
The enhanced representation is:
\begin{equation}
    \bar{\mathbf{z}}^{q}
    =
    \mathbf{z}^{q}
    +
    \alpha w_q(\mathbf{r}_{q}-\mathbf{z}^{q}),
    \quad
    w_q>\eta.
\end{equation}
For patches with insufficient morphology support, EasyAmplifier keeps the original representation unchanged:
\begin{equation}
    \bar{\mathbf{z}}^{q}
    =
    \mathbf{z}^{q},
    \quad
    w_q\leq\eta.
\end{equation}

\paragraph{Enhanced visual sequence.}
The final enhanced patch sequence is:
\begin{equation}
    \bar{\mathbf{Z}}
    =
    [\bar{\mathbf{z}}^{1},\bar{\mathbf{z}}^{2},\ldots,\bar{\mathbf{z}}^{P}].
\end{equation}
This sequence replaces the original visual embedding sequence before being passed to the subsequent medical VLM components. Since the enhancement uses only prototype retrieval, morphology-guided propagation, and residual injection, it requires no gradient-based optimization, no inference-time lesion annotations, and no modification of the pretrained model.

\section{Additional Implementation Details}
\label{B}
\subsection{Hyperparameter and Interface Calibration}
\label{app:hyperparam}

EasyLens is designed as a plug-and-play adapter for frozen medical VLMs. Its portability comes from separating the shared lesion-enhancement mechanism from lightweight interface calibration. The shared mechanism is unchanged across datasets and backbones: EasyTag selects lesion-relevant visual tokens, and EasyAmplifier applies residual lesion-aware enhancement to the selected tokens. The calibrated parameters only determine the operating point of this fixed mechanism.

\paragraph{Interface parameters.}
The main interface parameters are $\beta$, $k$, $s$, and $m$. The residual strength $\beta$ controls the magnitude of the EasyAmplifier update. The selected-token budget $k$ controls how many candidate visual tokens are retained by EasyTag. The local seed budget $s$ controls how many local visual peaks are used to initialize candidate lesion regions. The retrieval budget $m$ controls how many support tokens are retrieved from EasyBank.

Let $\mathbf{z}_i$ denote the original visual embedding of token $i$, and let $\mathcal{S}_k$ be the selected token set. EasyAmplifier updates selected tokens through a residual form:
\[
\bar{\mathbf{z}}_i =
\begin{cases}
\mathbf{z}_i + \beta \gamma_i(\mathbf{r}_i - \mathbf{z}_i), & i \in \mathcal{S}_k, \\
\mathbf{z}_i, & i \notin \mathcal{S}_k,
\end{cases}
\]
where $\mathbf{r}_i$ is the retrieved lesion-support reference and $\gamma_i$ is the calibrated lesion-support score from EasyTag and morphology-guided propagation. This residual formulation preserves the original visual representation while amplifying lesion-relevant evidence. In the method section, the residual coefficient is denoted by $\alpha$; in experiments, we use $\beta$ to denote its implementation value.

\paragraph{Operating regimes.}
We use three task-conditioned operating regimes. Counting tasks use a high-recall setting because missing a subtle lesion directly changes the answer. Selection tasks use a balanced precision--recall setting because excessive token amplification may introduce distractor regions. Report-generation tasks use a conservative setting because the decoder must preserve global context and language fluency.

For regular lesion report generation, the operating point is further calibrated according to lesion scale and morphology. Datasets with broader or more heterogeneous lesion regions require a larger token budget. For backbone adaptation, the same discrete grid is reused across VLMs, but the final operating point is calibrated to the visual-token geometry of each backbone. This calibration does not update any backbone parameters and does not change the EasyLens architecture.

\paragraph{Common settings.}
Unless otherwise specified, we use the common settings in Table~\ref{tab:app_common_settings}. All selected configurations are fixed before evaluation and shared by all test samples in the corresponding dataset--task or backbone--evaluation cell. No per-instance hyperparameter selection is used.

\begin{table}[h]
\centering
\caption{Common EasyLens settings used in the main experiments.}
\label{tab:app_common_settings}
\begin{tabular}{ll}
\toprule
Setting & Value \\
\midrule
Vision injection layer & 1 \\
Support mode & global top-$k$ \\
Seed mode & local peak top-$k$ \\
Seed peak kernel & 5 \\
Suppression radius & 1 \\
Score threshold & 0.0 \\
Default support score mode & \texttt{shape\_only} \\
Margin weight & 0.35 \\
Shape weight & 1.0 \\
Raw score weight & 1.0 \\
Gap score weight & 0.0 \\
Retrieval temperature & 0.1 \\
Max new tokens for count/select & 96 \\
Max new tokens for report generation & 160 \\
Report prompt mode & \texttt{benchmark\_raw} \\
Report probe window & first 3 decoding steps \\
\bottomrule
\end{tabular}
\end{table}

\subsection{Lesion-aware Report-generation Probe}
\label{app:report_probe}

For subtle-lesion report generation, lexical-overlap metrics alone may not fully capture whether the model recognizes the target lesion. A generated report can be fluent and globally plausible while still omitting the subtle abnormality. Therefore, in addition to generating free-form descriptions with the raw benchmark prompt, we use a fixed lesion-aware decoding probe to measure whether lesion-relevant tokens receive probability mass at the beginning of generation.

The probe is evaluation-only. It does not modify the prompt, decoder, visual encoder, EasyLens module, or generation procedure. For all report-generation experiments, we use the raw benchmark prompt, set the maximum generation length to 160 tokens, and compute lesion-token statistics over the first $T=3$ decoding steps.

Let $x_i$ be the input image and $q_i$ be the report-generation prompt for sample $i$. Given the enhanced visual representation $\bar{Z}_i$, the frozen VLM defines a next-token distribution:
\[
p_{i,t}(v)
=
p_{\theta}(v \mid q_i, \bar{Z}_i, y_{i,<t}),
\]
where $v$ is a vocabulary token and $t$ is the decoding step. For each sample, we define a strict lesion-token set $\mathcal{L}_i$ from the benchmark target description. This set contains tokens corresponding to the lesion category, morphology, or lesion-relevant clinical description.

At each decoding step, we compute the lesion-token probability mass:
\[
\ell_{i,t}
=
\sum_{v \in \mathcal{L}_i}
p_{i,t}(v).
\]
To avoid rewarding diffuse probability mass over a large vocabulary, we also use a strict top-$K$ version:
\[
\ell^{\mathrm{strict}}_{i,t}
=
\sum_{v \in \mathcal{L}_i \cap \mathrm{TopK}(p_{i,t})}
p_{i,t}(v),
\]
where $\mathrm{TopK}(p_{i,t})$ denotes the set of highest-probability vocabulary tokens at step $t$.

The final lesion-aware report-generation probe score is computed over the first three decoding steps:
\[
\mathrm{Probe}(i)
=
\frac{100}{T}
\sum_{t=1}^{T}
\ell^{\mathrm{strict}}_{i,t},
\quad T=3.
\]
The dataset-level score is the average over all report-generation samples:
\[
\mathrm{Probe}
=
\frac{1}{N}
\sum_{i=1}^{N}
\mathrm{Probe}(i).
\]

We focus on the first three decoding steps because early tokens usually determine the main clinical content of a generated report. If the model fails to assign probability mass to lesion-relevant tokens at this stage, later fluent continuation often cannot recover the missed subtle abnormality. This probe therefore measures lesion-awareness at the point where the model commits to the report content, while remaining independent of any training or inference-time modification.

\section{Additional Experimental Details}
\label{C}
\subsection{Dataset Details and Benchmark Construction}
\label{app:dataset_details}

\begin{wrapfigure}{r}{0.52\linewidth}
    \centering
    \includegraphics[width=0.98\linewidth]{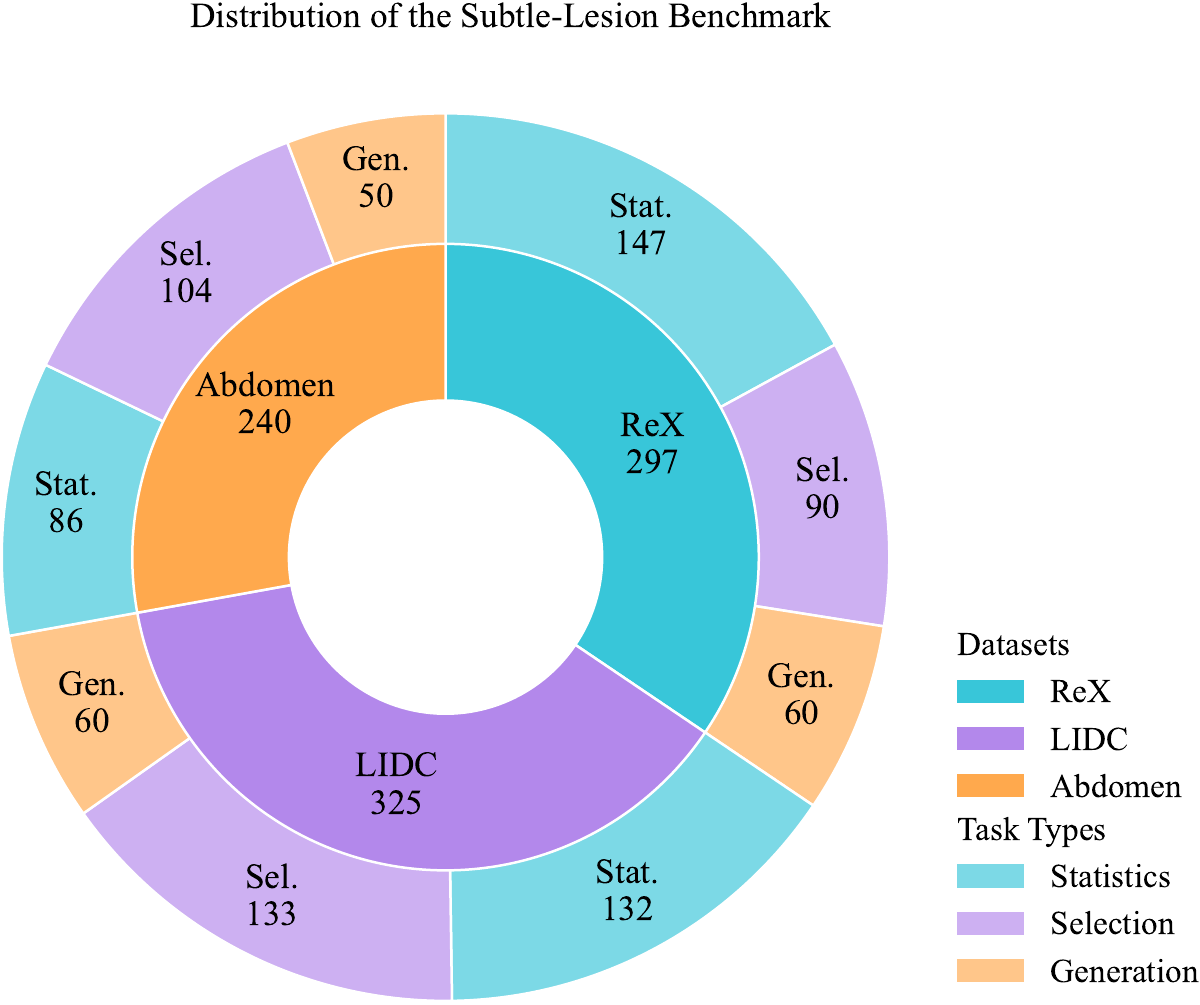}
    \caption{Dataset and task distribution of the subtle-lesion benchmark.}
    \label{fig:benchmark_distribution}
\end{wrapfigure}

We organize the evaluation datasets into two groups. The first group targets \emph{subtle lesions} and is built from ReXGroundingCT, LIDC-IDRI, and AbdomenAtlas 3.0 Mini, denoted as ReX, LIDC, and Abdomen, respectively. Following the terminology in the introduction, these cases focus on abnormalities whose visual evidence is spatially sparse, low-contrast, or weakly distinguishable from surrounding anatomical structures. The second group contains more salient or regular lesion datasets, including MIMIC-CXR, Kvasir-SEG, and BKAI-Polyp. These datasets are used to evaluate whether EasyLens preserves standard medical report-generation ability when pathological findings are visually more apparent.

For the subtle-lesion group, we construct a unified benchmark with three task types: regional statistics (Stat.), region selection (Sel.), and lesion-aware report generation (Gen.). Fig.~\ref{fig:benchmark_distribution} summarizes the distribution of the subtle-lesion benchmark across datasets and task types.

For Stat. and Sel., each image is divided into a $3\times3$ grid. Grid regions containing pathological patches are treated as positive regions. Stat. evaluates whether the model can correctly recover the number of positive regions, while Sel. evaluates whether it can identify which regions contain pathological evidence. These two tasks test complementary aspects of subtle-lesion perception: Stat. emphasizes regional evidence aggregation, whereas Sel. emphasizes spatial discrimination between lesion cues and visually similar anatomical background.

For subtle-lesion Gen., we evaluate whether subtle pathological evidence becomes more accessible during report generation. Instead of relying only on the final generated report, we measure the probability assigned to lesion-related tokens during decoding. This design is important because the enhanced visual embeddings may still be ignored by the decoder if the corresponding subtle-lesion tokens do not enter the final top-ranked generation path. Therefore, lesion-token probability provides a more direct signal of whether encoder-side enhancement actually increases the decoder's awareness of subtle pathological evidence.

For the salient or regular lesion group, we evaluate standard report generation on MIMIC-CXR, Kvasir-SEG, and BKAI-Polyp. Unlike the subtle-lesion Gen. task, this evaluation uses the final generated reports rather than lesion-token probabilities, because the goal is to test general report quality when the target findings are more visually apparent. We report BLEU-1, BLEU-4, METEOR, and ROUGE-L, which evaluate lexical overlap and sentence-level similarity between generated reports and reference reports. These results complement the subtle-lesion benchmark by testing whether EasyLens preserves general medical image interpretation and report-generation ability beyond subtle-lesion scenarios.

\subsection{Additional Evaluation on General and Salient-Lesion Report Generation}
\label{app:general_large}

\noindent\textbf{This section complements the main quantitative results by evaluating the benignness of EasyLens beyond subtle-lesion benchmarks.}
The main paper focuses on subtle-lesion perception, where EasyLens brings consistent gains across ReX, LIDC, and Abdomen. Here, we further evaluate whether the same enhancement mechanism affects standard report generation on general or more salient lesion cases. Table~\ref{tab:report_generation_comparison} reports results on MIMIC-CXR, Kvasir-SEG, and BKAI-Polyp.

\noindent\textbf{EasyLens preserves general chest X-ray reporting ability.}
On MIMIC-CXR, EasyLens remains highly comparable to its MedGemma1.5 backbone. It slightly improves BL-1 from 21.24 to 21.38, while BL-4, MTR, and RG-L remain nearly unchanged. This shows that enhancing subtle visual evidence does not disturb the global thoracic context or the original language-generation behavior of the frozen backbone.

\noindent\textbf{EasyLens also remains effective on salient lesion report generation.}
On Kvasir-SEG, EasyLens improves MedGemma1.5 from 9.56/15.87/15.39 to 10.30/16.55/16.18 on BL-1/MTR/RG-L. On BKAI-Polyp, it further improves the same backbone from 8.12/17.05/13.56 to 8.31/17.54/13.62. Since BL-4 is zero for all methods on the two polyp datasets, the comparison mainly relies on BL-1, MTR, and RG-L. These results support the conclusion in Sec.~\ref{sec:quant_eval}: EasyLens improves subtle-lesion perception without sacrificing broader medical report-generation ability.

\begin{table*}[t]
  \centering
  \small
  \setlength{\tabcolsep}{3.2pt}
  \renewcommand{\arraystretch}{1.18}

  \newcolumntype{C}[1]{>{\centering\arraybackslash}p{#1}}
  \newcolumntype{L}[1]{>{\raggedright\arraybackslash}p{#1}}

  \resizebox{\textwidth}{!}{
  \begin{tabular}{
    @{\hspace{3pt}}
    C{1.35cm}|L{1.15cm}|
    C{2.10cm}C{2.10cm}C{2.10cm}C{2.10cm}C{2.10cm}|C{1.15cm}
    @{\hspace{3pt}}
  }
    \hline\hline
    \textbf{Dataset} & \textbf{Metric}
    & \mbox{\textbf{LLaVA-Med}}
    & \textbf{RadFM}
    & \mbox{\textbf{Hulu-Med}}
    & \textbf{MedGemma}
    & \mbox{\textbf{MedGemma1.5}}
    & \textbf{EasyLens} \\
    \hline

     \multirow{4}{*}{\parbox[c]{1.35cm}{\centering\textbf{MIMIC-CXR}}}
    & BL-1   & 0.74 & 0.11 & 6.97 & 14.39 & 21.24 & \textbf{21.38} \\
    & BL-4   & 0.04 & 0.01 & 1.11 & 1.73 & \textbf{3.64} & 3.63 \\
    & MTR & 4.68 & 2.24 & 11.63 & 13.67 & \textbf{17.72} & 17.63 \\
    & RG-L   & 8.37 & 4.37 & 14.45 & 16.89 & \textbf{18.98} & 18.91 \\
    \hline

    \multirow{4}{*}{\parbox[c]{1.35cm}{\centering\textbf{Kvasir-SEG}}}
    & BL-1   & 11.49 & 1.60 & \textbf{16.14} & 13.99 & 9.56 & 10.30 \\
    & BL-4   & 0.00 & 0.00 & 0.00 & 0.00 & 0.00 & 0.00 \\
    & MTR & 22.69 & 1.31 & 12.49 & \textbf{25.38} & 15.87 & 16.55 \\
    & RG-L   & 8.61 & 2.41 & \textbf{21.40} & 15.59 & 15.39 & 16.18 \\
    \hline

    \multirow{4}{*}{\parbox[c]{1.35cm}{\centering\textbf{BKAI-Polyp}}}
    & BL-1   & 11.89 & 1.86 & 15.01 & \textbf{14.25} & 8.12 & 8.31 \\
    & BL-4   & 0.00 & 0.00 & 0.00 & 0.00 & 0.00 & 0.00 \\
    & MTR & 24.21 & 1.43 & 14.20 & \textbf{26.04} & 17.05 & 17.54 \\
    & RG-L   & 8.30 & 2.03 & \textbf{21.26} & 16.32 & 13.56 & 13.62 \\
    \hline\hline
  \end{tabular}
  }

  \vspace{6pt}
  \caption{Comparison of report generation performance across different medical VLMs on MIMIC-CXR, Kvasir-SEG, and BKAI-Polyp datasets.}
  \label{tab:report_generation_comparison}
  \vspace{-15pt}
\end{table*}

\subsection{Plug-and-Play Interface Calibration}
\label{app:interface_calibration}

\noindent\textbf{Plug-and-play deployment allows lightweight interface calibration without modifying the host VLM.}
EasyLens is attached to frozen medical VLMs as an inference-time adapter. Across all backbones, the VLM parameters are frozen, the decoding procedure is unchanged, and the EasyTag--EasyAmplifier architecture remains identical. The only model-dependent step is interface calibration, where a small number of scalar operating parameters are selected from a shared discrete grid.

\noindent\textbf{Interface calibration is necessary because different VLMs have different visual token geometries.}
Medical VLMs vary in embedding scale, lesion-background separability, and the spatial distribution of lesion evidence. A residual strength that is suitable for one encoder may be too weak for another encoder with lower lesion contrast, or too strong for an encoder whose lesion tokens are already sharply localized. Similarly, some backbones concentrate lesion evidence in a few patches, while others distribute it across more visual tokens. Therefore, these calibrated parameters should be interpreted as adapter-interface parameters, not as model-specific learned components.

\noindent\textbf{The calibrated parameters control only the strength and coverage of the same shared mechanism.}
The main parameters are $\beta$, $k$, $s$, and $m$. The residual strength $\beta$ controls how strongly selected visual tokens are moved toward lesion-support directions. The selector budget $k$ controls how many candidate tokens are retained. The seed budget $s$ determines how many local visual peaks initialize candidate lesion regions. The retrieval budget $m$ controls how many support tokens are retrieved from the EasyBank. These parameters are low-dimensional, interpretable, and selected from a small discrete grid.

\noindent\textbf{EasyLens does not tune parameters per test example.}
For each dataset--task or backbone--task setting, the operating point is fixed before evaluation and then applied to all test samples. Thus, the calibration is an interface-level deployment choice rather than instance-level optimization. No backbone parameters are updated, no model-specific EasyLens architecture is introduced, and no task-specific instruction tuning is performed.

\noindent\textbf{Task-conditioned regimes reflect different precision--recall requirements.}
For status or counting-oriented tasks, EasyLens uses a high-recall regime because missing a subtle lesion may directly change the answer. For lesion selection, EasyLens uses a balanced regime that preserves enough candidate regions while avoiding excessive distractor amplification. For report generation, EasyLens uses a conservative-generation regime with stable token coverage and moderate residual strength, so that lesion awareness is improved without destabilizing global context or language fluency.

\noindent\textbf{Regular lesion report generation uses the same calibration principle.}
MIMIC-CXR adopts a conservative setting because reports require broad thoracic context. Kvasir-SEG uses a moderate setting because the lesion is localized and visually salient. BKAI-Polyp uses a larger coverage setting because polyp appearance and boundary morphology are more heterogeneous. These choices follow the same interface-calibration principle used for subtle-lesion tasks.

\subsection{Common Implementation Settings}
\label{app:common_settings}

\noindent\textbf{Unless otherwise specified, all experiments share the same EasyLens implementation settings.}
EasyLens is applied to the first vision layer. We use global top-$k$ support retrieval, local-peak seed initialization, a seed peak kernel size of 5, suppression radius 1, margin weight 0.35, shape weight 1.0, score threshold 0.0, and retrieval temperature 0.1. The default support scoring mode is \texttt{shape\_only}. For VQA-style tasks, the maximum number of generated tokens is 96. For report-generation tasks, the maximum number of generated tokens is 160, and the lesion-token scoring window uses three decoding steps.

\begin{table*}[t]
  \centering
  \small
  \setlength{\tabcolsep}{2.8pt}
  \renewcommand{\arraystretch}{1.16}
    \begin{tabularx}{\textwidth}{
        @{\hspace{5pt}}
        l|
        *{3}{>{\centering\arraybackslash}X}|
        *{3}{>{\centering\arraybackslash}X}|
        *{3}{>{\centering\arraybackslash}X}|
        >{\centering\arraybackslash}X
        @{}
    }
    \hline\hline
    \multirow{2}{*}{\textbf{Layer}} &
    \multicolumn{3}{c|}{\textbf{ReX}} &
    \multicolumn{3}{c|}{\textbf{LIDC}} &
    \multicolumn{3}{c|}{\textbf{Abdomen}} &
    \multirow{2}{*}{\textbf{Macro}} \\
    \cline{2-10}
    & \textbf{Stat.} & \textbf{Sel.} & \textbf{Gen.}
    & \textbf{Stat.} & \textbf{Sel.} & \textbf{Gen.}
    & \textbf{Stat.} & \textbf{Sel.} & \textbf{Gen.}
    & \\
    \hline

    1
    & \textbf{66.67} & \textbf{31.11} & \textbf{5.15}
    & \textbf{30.30} & \textbf{36.09} & \textbf{45.86}
    & \textbf{52.33} & \textbf{55.77} & \textbf{40.67}
    & \textbf{40.44} \\

    6
    & 48.98 & 24.44 & 4.74
    & 24.24 & 30.08 & 44.28
    & 45.35 & 50.00 & 38.77
    & 34.54 \\

    11
    & 46.94 & 25.56 & 5.04
    & 22.73 & 30.08 & 41.36
    & 45.35 & 49.04 & 37.98
    & 33.79 \\

    16
    & 44.22 & 24.44 & 4.51
    & 18.18 & 30.83 & 41.51
    & 45.35 & 51.92 & 37.34
    & 33.14 \\

    21
    & 46.26 & 23.33 & 4.41
    & 17.42 & 31.58 & 43.09
    & 48.84 & 50.96 & 37.89
    & 33.75 \\

    26
    & 57.14 & 25.56 & 5.02
    & 21.21 & 30.83 & 40.75
    & 46.51 & 47.12 & 38.56
    & 34.74 \\
    \hline\hline
  \end{tabularx}

  \vspace{6pt}
  \caption{Ablation of the vision layer used by EasyLens on ReX, LIDC, and Abdomen. Early injection achieves the best overall performance.}
  \label{tab:layer_ablation}
\end{table*}

\begin{table*}
  \centering
  \small
  \setlength{\tabcolsep}{2.8pt}
  \renewcommand{\arraystretch}{1.16}
    \begin{tabularx}{\textwidth}{
        @{\hspace{5pt}}
        l|
        *{3}{>{\centering\arraybackslash}X}|
        *{3}{>{\centering\arraybackslash}X}|
        *{3}{>{\centering\arraybackslash}X}
        @{}
    }
    \hline\hline
    \multirow{2}{*}{\textbf{Setting}} &
    \multicolumn{3}{c|}{\textbf{ReX}} &
    \multicolumn{3}{c|}{\textbf{LIDC}} &
    \multicolumn{3}{c}{\textbf{Abdomen}} \\
    \cline{2-10}
    & \textbf{Stat.} & \textbf{Sel.} & \textbf{Gen.}
    & \textbf{Stat.} & \textbf{Sel.} & \textbf{Gen.}
    & \textbf{Stat.} & \textbf{Sel.} & \textbf{Gen.} \\
    \hline

    topk\_32\_seed4
    & 46.26 & 26.67 & \textbf{5.17}
    & \textbf{26.52} & 30.08 & 42.40
    & 47.67 & 49.04 & \textbf{39.79} \\

    topk\_64\_seed8
    & 45.58 & 26.67 & 3.93
    & 21.21 & 30.08 & 42.52
    & \textbf{52.33} & 48.08 & 37.96 \\

    topk\_128\_seed16
    & 54.42 & 31.11 & 5.15
    & \textbf{26.52} & 29.32 & 43.16
    & 48.84 & \textbf{52.88} & 39.76 \\

    topk\_256\_seed32
    & 66.67 & 28.89 & 4.04
    & 25.76 & \textbf{31.58} & \textbf{45.46}
    & \textbf{52.33} & \textbf{52.88} & 38.18 \\

    topk\_512\_seed48
    & \textbf{71.43} & \textbf{32.22} & 3.86
    & 25.00 & 29.32 & 44.76
    & 48.84 & 51.92 & 38.87 \\
    \hline\hline
  \end{tabularx}

  \vspace{6pt}
  \caption{Ablation of the EasyTag candidate token budget $k$ and seed budget $s$ on ReX, LIDC, and Abdomen. Different tasks prefer different operating points due to distinct precision--recall requirements.}
  \label{tab:topk_seed_results}
\end{table*}

\begin{figure}
    \centering
    \begin{minipage}{0.58\linewidth}
        \centering
        \includegraphics[width=\linewidth]{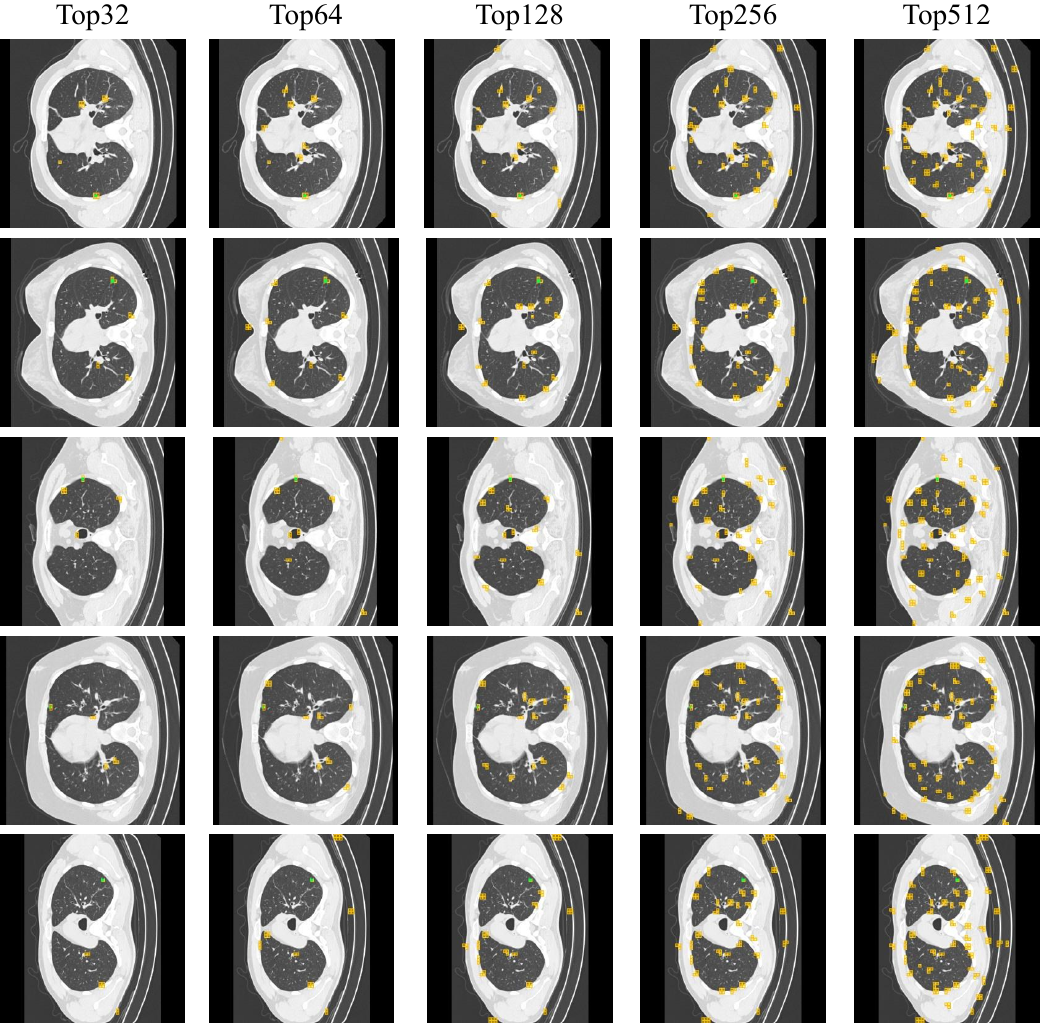}
        \caption{Case study for patch selection in different topk settings}
        \label{fig:figure_cs2}
    \end{minipage}
\end{figure}

\subsection{Additional Interface Ablations}
\label{app:ablation_detail}

\noindent\textbf{These ablations complement the main component ablation in Sec.~\ref{sec:ablation}.}
The main paper verifies the effectiveness of EasyTag and EasyAmplifier. Here, we further analyze how the same EasyTag--EasyAmplifier mechanism should be interfaced with frozen VLMs, including the visual injection position and the candidate token budget. These experiments do not introduce additional task-specific modules; they only study the operating interface of the fixed EasyLens design.

\subsubsection{Visual Injection Position}
\label{app:injection_position}

\noindent\textbf{Early visual injection is the most effective for preserving subtle lesion evidence.}
Table~\ref{tab:layer_ablation} compares injecting EasyLens into different vision layers. Injecting at Layer 1 achieves the best results across all nine subtle-lesion metrics and the highest macro score of 40.44. Moving the injection to deeper layers consistently reduces the macro score, e.g., 34.54 at Layer 6, 33.79 at Layer 11, and 33.14 at Layer 16. Although Layer 26 partially recovers the ReX Stat. score, it remains clearly weaker than Layer 1 on the overall macro score and on LIDC/Abdomen.

\noindent\textbf{This supports using EasyLens as an early visual-interface adapter.}
Subtle lesions are low-contrast and spatially localized. If enhancement is applied too late, patch-level lesion evidence may already be compressed by the vision encoder and multimodal alignment layers. Early residual injection amplifies lesion-relevant visual tokens before this compression, while still allowing the frozen VLM to perform its original downstream reasoning.

\noindent\textbf{The candidate token budget reflects a precision--recall trade-off.}
Table~\ref{tab:topk_seed_results} varies the EasyTag candidate budget $k$ together with the seed budget $s$. No single budget dominates all datasets and metrics. On ReX, a larger budget improves Stat. and Sel., while Gen. is better with a compact budget. On LIDC, a larger budget benefits selection and generation, with $k=256$ achieving the best Sel. and Gen. On Abdomen, selection prefers a moderate-to-large budget, while generation again benefits from a smaller candidate set.

\noindent\textbf{This behavior is consistent with task-conditioned calibration.}
Status and selection tasks often require high recall over candidate lesion regions, especially when multiple subtle abnormalities may be present. Report generation, however, must preserve lesion evidence without introducing excessive background distractors. When $k$ is too small, the selector may miss weak lesion evidence; when $k$ is too large, additional candidates may dilute the lesion signal. The top-$k$ ablation therefore supports the use of a fixed EasyLens mechanism with lightweight interface calibration rather than a one-size-fits-all operating point.

\subsubsection{Qualitative Analysis of Candidate Token Budget}
\label{app:topk_case}

\noindent\textbf{The topk case study provides a visual explanation for the budget ablation.}
Figure~\ref{fig:figure_cs2} visualizes how different candidate budgets affect lesion evidence selection. A small budget tends to focus on the most confident local responses, which can reduce distractors but may miss weak or spatially scattered lesions. A moderate budget preserves more lesion-related alternatives and is therefore more suitable for selection-oriented tasks. An excessively large budget may introduce background structures with similar local appearance, diluting the lesion signal during generation.

\noindent\textbf{This qualitative behavior matches the quantitative trend in Table~\ref{tab:topk_seed_results}.}
The visualization shows that EasyLens does not rely on arbitrary token amplification. Instead, the candidate budget controls the amount of visual evidence exposed to the frozen VLM. Proper calibration balances lesion recall and background suppression, which is crucial for subtle-lesion grounding and faithful report generation.

\section{Limitations}
\label{app:limitations}

EasyLens has several limitations. First, although it is training-free at inference time, it relies on an offline prototype bank constructed from reference images and lesion masks. Its effectiveness may therefore depend on the coverage of the prototype bank, especially when target lesions differ in anatomy, scale, or imaging appearance from the reference set. Second, the current implementation uses fixed interface parameters for each dataset--task setting, including the token budget and residual strength. This simple calibration strategy may not be optimal for all cases, particularly when lesion size or visual contrast varies substantially. Third, our experiments are conducted on a limited set of medical VLM backbones and lesion-oriented benchmarks. Broader validation across additional imaging modalities, institutions, and clinical tasks is still needed to assess the generality of the proposed approach.

\section{Licenses and Terms for Existing Assets}
\label{app:asset_licenses}

This work uses existing public or credentialed medical datasets and publicly released model checkpoints only for research evaluation. We do not claim ownership of any third-party assets. We cite the original creators of the datasets, models, and codebases used in our experiments, and we use each asset according to its corresponding license, data-use agreement, or access policy. We do not redistribute third-party medical images, reports, masks, or model weights. Users who reproduce our experiments should obtain the original assets from their official sources and comply with the corresponding licenses and access requirements.

\begin{table}[h]
\centering
\small
\caption{Existing datasets and model assets used in this work. For assets with restricted or gated access, users must obtain access from the original provider.}
\label{tab:asset_licenses}
\begin{tabular}{p{0.20\linewidth} p{0.28\linewidth} p{0.45\linewidth}}
\toprule
\textbf{Asset} & \textbf{Role in this work} & \textbf{License or access terms} \\
\midrule
ReXGroundingCT & Subtle-lesion CT evaluation and prototype construction & Released under CC BY-NC-SA 4.0 with gated access requiring acceptance of dataset conditions. We use it only for non-commercial research evaluation and do not redistribute the data. \\

LIDC-IDRI & Lung CT subtle-lesion evaluation and prototype construction & Distributed through TCIA. We follow the TCIA Data Usage Policy and the Creative Commons Attribution 3.0 Unported license terms associated with the collection. We cite the original dataset and do not redistribute the data. \\

AbdomenAtlas 3.0 Mini & Abdominal CT subtle-lesion evaluation and prototype construction & Released under CC BY-NC-SA 4.0. We use it for non-commercial research evaluation and do not redistribute the data. \\

MIMIC-CXR & Chest X-ray report-generation evaluation & Accessed through PhysioNet under the PhysioNet Credentialed Health Data License 1.5.0 and the corresponding Data Use Agreement. Access requires credentialed-user approval and required training. We do not share the data, attempt re-identification, or redistribute any images or reports. \\

Kvasir-SEG & Polyp report-generation evaluation & The official terms restrict use to research and educational purposes, and commercial use requires prior written permission. We cite the dataset paper and do not redistribute the dataset. \\

BKAI-Polyp / BKAI-IGH NeoPolyp & Polyp report-generation evaluation & Accessed as Kaggle competition data subject to the original competition rules. We use it only for research evaluation and do not redistribute the dataset. \\

LLaVA-Med & Frozen medical VLM backbone and baseline & The official release states that the data, code, and model checkpoints are intended for research use only, with non-commercial restrictions and additional terms inherited from the underlying LLaMA, Vicuna, and GPT-4 resources. It is not intended for clinical care or clinical decision making. \\

RadFM & Frozen medical VLM backbone and baseline & The official code repository is released under the MIT License. The public checkpoint page used in our experiments does not specify a separate model-card license; therefore, we use the checkpoint only for research evaluation and do not redistribute the weights. \\

Lingshu & Frozen medical VLM backbone and baseline & The official Hugging Face model card lists the MIT License. We use the released checkpoint only for research evaluation and not for clinical deployment. \\

Hulu-Med & Frozen medical VLM backbone and baseline & The official repository is released under the Apache License 2.0. We use the released checkpoint and code only for research evaluation and not for clinical deployment. \\

MedGemma and MedGemma1.5 & Frozen medical VLM backbone and baseline & The model weights are governed by the Health AI Developer Foundations terms of use. The accompanying repository/tutorial code is released under Apache License 2.0. We use MedGemma models only as frozen research backbones and do not treat them as clinical-grade diagnostic systems. \\

MedKLIP & Encoder-enhancement baseline & We cite the official paper and repository. We did not identify a separate explicit license for the released repository or checkpoint at the time of writing; therefore, we use it only for research comparison and do not redistribute its code or weights. \\

KAD & Encoder-enhancement baseline & The official code repository is released under the MIT License. Any associated model or data resources are used according to their original access terms, and we do not redistribute third-party assets. \\
\bottomrule
\end{tabular}
\end{table}

Any supplementary material released with this submission contains only our implementation, configuration files, and evaluation scripts. It does not include protected health information, third-party medical images, clinical reports, segmentation masks, or third-party model weights. Reproduction of the reported experiments requires users to independently obtain the relevant datasets and model checkpoints from their original providers and to comply with the corresponding licenses, data-use agreements, and citation requirements.

\clearpage
\section*{NeurIPS Paper Checklist}

\begin{enumerate}

\item {\bf Claims}
    \item[] Question: Do the main claims made in the abstract and introduction accurately reflect the paper's contributions and scope?
    \item[] Answer: \answerYes{}.
    \item[] Justification: The abstract and Introduction clearly state that EasyLens is a training-free plug-and-play subtle-lesion representation amplifier for frozen medical VLMs. The claimed contributions are limited to EasyBank, EasyTag, and EasyAmplifier, and the experimental claims are supported on ReXGroundingCT, LIDC-IDRI, AbdomenAtlas 3.0 Mini, MIMIC-CXR, Kvasir-SEG, and BKAI-Polyp with multiple frozen medical VLM backbones.

\item {\bf Limitations}
    \item[] Question: Does the paper discuss the limitations of the work performed by the authors?
    \item[] Answer: \answerYes{}.
    \item[] Justification: Appendix~\ref{app:limitations} discusses limitations related to reliance on prototype construction masks, dataset and modality coverage, sensitivity to interface calibration, possible false-positive amplification, and the need for prospective clinical validation before deployment.

\item {\bf Theory assumptions and proofs}
    \item[] Question: For each theoretical result, does the paper provide the full set of assumptions and a complete (and correct) proof?
    \item[] Answer: \answerNA{}.
    \item[] Justification: The paper does not present theorem-level theoretical results or formal proofs. The mathematical formulations in Section 3 and Appendix A define the EasyBank, EasyTag, and EasyAmplifier procedures rather than proving theoretical guarantees.

    \item {\bf Experimental result reproducibility}
    \item[] Question: Does the paper fully disclose all the information needed to reproduce the main experimental results of the paper to the extent that it affects the main claims and/or conclusions of the paper (regardless of whether the code and data are provided or not)?
    \item[] Answer: \answerYes{}.
    \item[] Justification: Sections 3 and 4 describe the proposed algorithm, frozen-backbone setting, evaluation tasks, metrics, and compared backbones. Appendix~\ref{app:method_details} provides detailed method definitions, Appendix~\ref{B} specifies interface calibration and common implementation settings, and Appendix~\ref{C} describes dataset/task construction and additional evaluations.

\item {\bf Open access to data and code}
    \item[] Question: Does the paper provide open access to the data and code, with sufficient instructions to faithfully reproduce the main experimental results, as described in supplemental material?
    \item[] Answer: \answerYes{}.
    \item[] Justification: The submission includes anonymized code and reproduction instructions in the supplementary material. The experiments use existing public or credentialed medical datasets, and the paper describes the preprocessing, task construction, evaluation metrics, and EasyLens implementation settings needed to reproduce the main results.

\item {\bf Experimental setting/details}
    \item[] Question: Does the paper specify all the training and test details (e.g., data splits, hyperparameters, how they were chosen, type of optimizer) necessary to understand the results?
    \item[] Answer: \answerYes{}.
    \item[] Justification: Sections 4.1 and 4.2 describe the datasets, task types, metrics, frozen VLM backbones, raw benchmark prompts, and generation length. Appendix~\ref{app:hyperparam} and Table~\ref{tab:app_common_settings} specify the EasyLens interface parameters and common settings, including injection layer, support retrieval mode, seed initialization, score threshold, retrieval temperature, and maximum generated tokens. Since EasyLens is training-free, optimizer and training schedule details are not applicable.

\item {\bf Experiment statistical significance}
    \item[] Question: Does the paper report error bars suitably and correctly defined or other appropriate information about the statistical significance of the experiments?
    \item[] Answer: \answerYes{}.
    \item[] Justification: The temperature is set to 0 and the results are repeatable.

\item {\bf Experiments compute resources}
    \item[] Question: For each experiment, does the paper provide sufficient information on the computer resources (type of compute workers, memory, time of execution) needed to reproduce the experiments?
    \item[] Answer: \answerYes{}.
    \item[] Justification: All experiments all running on A6000 with 48GB memory,and table~\ref{tab:comparison_wo_w} reports the average inference time.
    
\item {\bf Code of ethics}
    \item[] Question: Does the research conducted in the paper conform, in every respect, with the NeurIPS Code of Ethics \url{https://neurips.cc/public/EthicsGuidelines}?
    \item[] Answer: \answerYes{}.
    \item[] Justification: The research uses existing medical image datasets and evaluates a frozen-model inference-time adapter. It does not collect new patient data or conduct new human-subject experiments. The method is intended for research on medical VLM sensitivity and should not be used as an autonomous clinical diagnostic system without appropriate validation and oversight.

\item {\bf Broader impacts}
    \item[] Question: Does the paper discuss both potential positive societal impacts and negative societal impacts of the work performed?
    \item[] Answer: \answerYes{}.
    \item[] Justification: We discuss the potential benefit of improving sensitivity to clinically important subtle lesions, which may support more reliable medical image interpretation. We also note that EasyLens is not a standalone diagnostic system and should be used with clinician oversight, with attention to validation across populations, institutions, and privacy-sensitive medical data handling.
    
\item {\bf Safeguards}
    \item[] Question: Does the paper describe safeguards that have been put in place for responsible release of data or models that have a high risk for misuse (e.g., pre-trained language models, image generators, or scraped datasets)?
    \item[] Answer: \answerNA{}.
    \item[] Justification: The paper does not release a new high-risk foundation model, scraped dataset, image generator, or autonomous clinical system. EasyLens is an inference-time adapter evaluated on existing medical VLMs and datasets.

\item {\bf Licenses for existing assets}
    \item[] Question: Are the creators or original owners of assets (e.g., code, data, models), used in the paper, properly credited and are the license and terms of use explicitly mentioned and properly respected?
    \item[] Answer: \answerYes{}.
    \item[] Justification: The paper cites the original sources of the datasets and model backbones used in the experiments, and the supplementary material lists the corresponding versions, licenses, and access terms in Appendix~\ref{app:asset_licenses}. All existing assets are used according to their stated research-use or credentialed-access policies.

\item {\bf New assets}
    \item[] Question: Are new assets introduced in the paper well documented and is the documentation provided alongside the assets?
    \item[] Answer: \answerYes{}.
    \item[] Justification: The paper introduces a unified subtle-lesion benchmark derived from existing datasets and documents the task construction, dataset/task distributions, evaluation metrics, and usage protocol in Appendix C. The released asset is anonymized and accompanied by documentation for reproduction.

\item {\bf Crowdsourcing and research with human subjects}
    \item[] Question: For crowdsourcing experiments and research with human subjects, does the paper include the full text of instructions given to participants and screenshots, if applicable, as well as details about compensation (if any)? 
    \item[] Answer: \answerNA{}.
    \item[] Justification: The work does not involve crowdsourcing, user studies, or new human-subject data collection.

\item {\bf Institutional review board (IRB) approvals or equivalent for research with human subjects}
    \item[] Question: Does the paper describe potential risks incurred by study participants, whether such risks were disclosed to the subjects, and whether Institutional Review Board (IRB) approvals (or an equivalent approval/review based on the requirements of your country or institution) were obtained?
    \item[] Answer: \answerNA{}.
    \item[] Justification: The study does not collect new human-subject data and uses existing medical imaging datasets under their corresponding access and usage policies. No new intervention, recruitment, or participant interaction is involved.

\item {\bf Declaration of LLM usage}
    \item[] Question: Does the paper describe the usage of LLMs if it is an important, original, or non-standard component of the core methods in this research? Note that if the LLM is used only for writing, editing, or formatting purposes and does \emph{not} impact the core methodology, scientific rigor, or originality of the research, declaration is not required.
    \item[] Answer: \answerNA{}.
    \item[] Justification: LLMs were used only for writing, editing, and formatting support and do not affect the core methodology, scientific rigor, or originality of the research.

\end{enumerate}

\end{document}